\documentclass[letterpaper]{article} 
\usepackage[]{aaai2026}  
\usepackage{times}  
\usepackage{helvet}  
\usepackage{courier}  
\usepackage[hyphens]{url}  
\usepackage{graphicx} 
\usepackage{natbib}  
\usepackage{caption} 
\usepackage{algorithm}
\usepackage{algorithmic}
\nocopyright
\usepackage{newfloat}
\usepackage{listings}
\DeclareCaptionStyle{ruled}{labelfont=normalfont,labelsep=colon,strut=off} 
\floatstyle{ruled}
\newfloat{listing}{tb}{lst}{}
\floatname{listing}{Listing}
\title{Distribution Shift Aware Neural Tabular Learning}
\author {
    Wangyang Ying\textsuperscript{\rm 1},
    Nanxu Gong\textsuperscript{\rm 1},
    Dongjie Wang\textsuperscript{\rm 2},
    Xinyuan Wang\textsuperscript{\rm 1},
    Arun Vignesh Malarkkan\textsuperscript{\rm 1},
    Vivek Gupta\textsuperscript{\rm 1},
    Chandan K. Reddy\textsuperscript{\rm 3},
    Yanjie Fu\textsuperscript{\rm 1}
}
\affiliations {
    \textsuperscript{\rm 1}Arizona State University\\
    \textsuperscript{\rm 2}The University of Kansas\\
    \textsuperscript{\rm 3}Virginia Tech\\
    \{wying4, nanxugong, xwang735, arun.malarkkan, vgupt140, yanjie.fu\}@asu.edu, wangdongjie@ku.edu, reddy@cs.vt.edu
}

\usepackage{soul}
\usepackage{subfigure}
\usepackage{booktabs} 
\usepackage{amsmath}
\usepackage{tikz}
\def\model{SAFT}
\newcommand{\blackcircnum}[1]{%
  \tikz[baseline={(char.base)}]{
    \node[shape=circle,fill=black,inner sep=1pt,
          text=white,font=\bfseries\footnotesize] (char) {#1};}}
          
\def\BibTeX{{\rm B\kern-.05em{\sc i\kern-.025em b}\kern-.08em
    T\kern-.1667em\lower.7ex\hbox{E}\kern-.125emX}}

\begin{document}

\maketitle

\begin{abstract}
Tabular learning transforms raw features into optimized spaces for downstream tasks, but its effectiveness deteriorates under distribution shifts between training and testing data. We formalize this challenge as the Distribution Shift Tabular Learning (DSTL) problem and propose a novel Shift-Aware Feature Transformation (SAFT) framework to address it. SAFT reframes tabular learning from a discrete search task into a continuous representation-generation paradigm, enabling differentiable optimization over transformed feature sets. SAFT integrates three mechanisms to ensure robustness: (i) shift-resistant representation via embedding decorrelation and sample reweighting, (ii) flatness-aware generation through suboptimal embedding averaging, and (iii) normalization-based alignment between training and test distributions. Extensive experiments show that SAFT consistently outperforms prior tabular learning methods in terms of robustness, effectiveness, and generalization ability under diverse real-world distribution shifts.
\end{abstract}

\section{Introduction}
\begin{figure}[t]
\centering
\includegraphics[width=\linewidth]{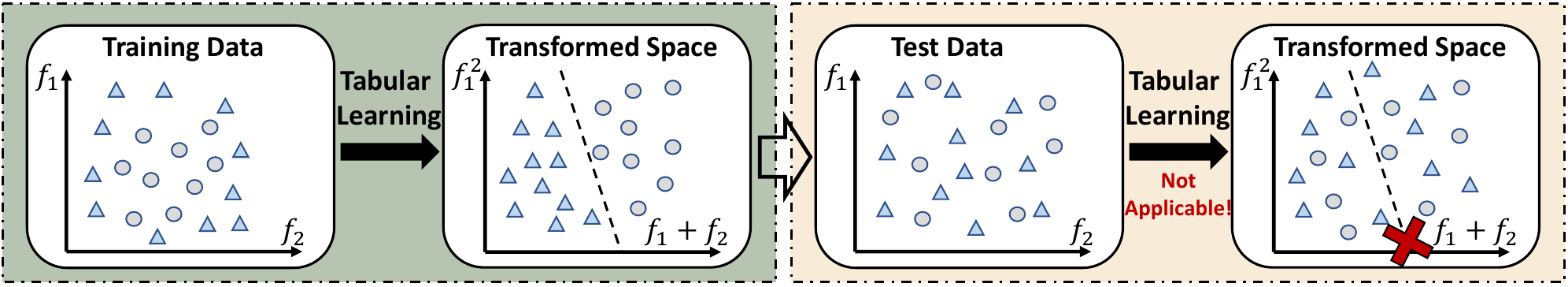}
\vspace{-0.3cm}
\caption{DSTL shows how distribution shifts between training and test sets impact tabular learning. The transformation applied during training fails to generalize in the test domain.}
\vspace{-0.5cm}
\label{motivation}
\end{figure}

Deep learning relies on large models and expensive GPUs. Data-Centric AI offers a cost-effective alternative by improving data quality to boost performance~\cite{zha2023data}.
Tabular data is prevalent in both academia and industry, where designing effective features is critical~\cite{kusiak2001feature, ying2023self}.
A key challenge arises when feature distributions shift between training and testing, even within the same domain, violating the common I.I.D. assumption.
As a result, features learned during training can fail at test time (see \textbf{Figure \ref{motivation}}).
We define this as the Distribution Shift Tabular Learning (DSTL) problem, which is crucial for improving generalization and robustness in real-world tabular data.

Related works solve DSTL only partially. \blackcircnum{1} DSTL is related to feature transformation (e.g., $\{f_1,f_2\}\rightarrow\{f_1+f_2\}$)~\cite{ying2025survey}. For instance, humans can manually transform a feature set with domain knowledge. Machine-assisted methods include principal components analysis (PCA)~\cite{mackiewicz1993principal}, exhaustive-expansion-reduction approaches~\cite{kanter2015deep, khurana2016cognito}, iterative-feedback-improvement approaches~\cite{khurana2018feature, ying2023self, ying2024topology}. 
However, manual methods are time-consuming and incomplete. Among machine-assisted methods, PCA is only based on a straight linear feature correlation assumption, ignoring non-linear feature interaction, and PCA only reduces dimensionality instead of increasing dimensionality; other methods mostly rely on discrete search formulation with critical limitations: 1) large search space and time costly when enumerating combinations, 2) lacks generalization and robustness against imperfect data, such as distribution shifts.
\blackcircnum{2} DSTL is related to anti-shift learning, for instance, distribution alignment~\cite{fan2024addressing,kim2021reversible,hu2023boosting}, parameter isolation~\cite{zhang2023continual}, adaptive experience replay~\cite{li2024adaer}.  
A key research gap remains: how can shift awareness be integrated into learning-based formulations that go beyond traditional discrete search?

There are two challenges to bridge the gap between feature transformation and shift awareness in tabular learning: 
1) \ul{From search-based formulation to learning-based formulation.} 
Existing studies view feature transformation as a search task for the best feature set among large, discrete feature combinations. The first challenge seeks to answer: what paradigm allows us to frame feature transformation as a learning problem - within a continuous space, guided by directional gradients, evaluated by tangible objectives, and measured using discriminative or generative models?

\noindent 2) \ul{From shift-sensitive to shift-robust.}
Once the feature transformation learning paradigm is proposed, the second challenge intends to answer: what mechanisms can integrate shift awareness into such a learning paradigm? 

\textbf{Insight: Shift-aware Representation-Generation.}
We reformulate the DSTL problem as an optimization task in continuous space rather than a discrete search task. Imagine each discrete combination (e.g., the transformed feature set $\{f_1+f_2,f_3*f_4\}$) is a point in a high-dimensional space. By learning a continuous embedding of these combinations, we can reason over their relationships and apply gradient-based optimization to search for better combinations.
Although this representation-generation paradigm is common in domains like computer vision, its application to classical feature engineering introduces a new opportunity to integrate shift awareness into representation, generation, and pre/post-processing.

\textbf{Summary of Proposed Solution.}
We adopt a representation generation-based deep tabular learning framework. In the representation step, an encoder is trained to embed transformed feature sets into vectors; an evaluator is optimized to predict performance from embeddings, and a decoder is learned to reconstruct the feature set from the embedding. The generation step then applies gradient ascent to identify the optimal embedding and decodes it into the best-transformed feature set.
To address the distribution shift, we develop three mechanisms in response to shift-resistant representation, flatness-aware generation, and shift-aligned pre and post-processing: (i) Connecting embedding dimension decorrelation with sample reweighing can incorporate shift resistance into embedding. (ii) Leveraging flatness in gradient-based optimal embedding search can ensure that the performance of a transformed feature set embedding does not decrease significantly in a neighborhood around the maximum, thus mitigating distribution shift. (iii) Performing normalization in pre-processing and denormalization in post-processing can align distribution gaps between training and testing data. 

\textbf{Our contributions} are:
1) We propose a representation-generation framework that enables continuous modeling of discrete feature transformations, addressing distribution shift in DSTL tasks.
2) We introduce key shift-resilient mechanisms (i.e., feature set representation, flatness-aware generation, and normalization) to enhance robustness under domain shifts.
3) Extensive experiments on benchmark datasets demonstrate the proposed method's superior performance under distribution shift.

\section{Problem Statement}
\noindent\textbf{The DSTL Problem.}
Given a training dataset and a test dataset that share the same set of features, and considering the existence of distribution shifts between training and test samples, we aim to learn a distribution shift-resilient tabular learning model that transforms the original datasets into a new representation space. The goal is to: 1) improve the performance of an ML task (e.g., regression, classification) and 2) ensure robustness to distribution shift. 
We adopt the perspective that knowledge of which feature combinations enhance data utility can be learned from past feature transformation–performance pairs. Based on this view, we formulate the DSTL problem as a shift-aware generative tabular learning task involving the following key concepts:

\noindent\textbf{1) Feature Cross.} We apply operations (e.g., $+$, $-$, $*$, $sin$($\cdot$)) 
to cross original features (e.g., $f_1+f_2$). 

\noindent\textbf{2) Feature Cross Sequence.} We convert feature crosses into postfix expressions and represent them as token sequences to reduce token redundancy and prevent invalid transformations. We concatenate multiple feature crosses using a separation token, and we prepend a start token and append an end token to the full sequence. For example, the two feature crosses $f_1+f_1/f_3-f_2,\sqrt{f_1}$ are represented by: $<\text{sos}>f_2f_1f_3/f_3-+<\text{sep}>f_2\sqrt{.}<\text{eos}>$. 

\noindent\textbf{3) Transformed Feature Set.} A feature cross sequence defines a set of transformation rules. By applying each rule to the original features, we generate a collection of new features, which together form the transformed feature set.

We collect various feature cross sequences along with their corresponding performances and transformed feature sets as \textbf{observed values} from the training dataset. DSTL encodes these observed values into an embedding space and generates a new feature cross sequence. Using the transformation rules defined by the new feature cross sequence, we convert training and testing data points into a new space, thereby improving downstream performance.

\begin{figure}[h]
\centering
\includegraphics[width=1\linewidth]{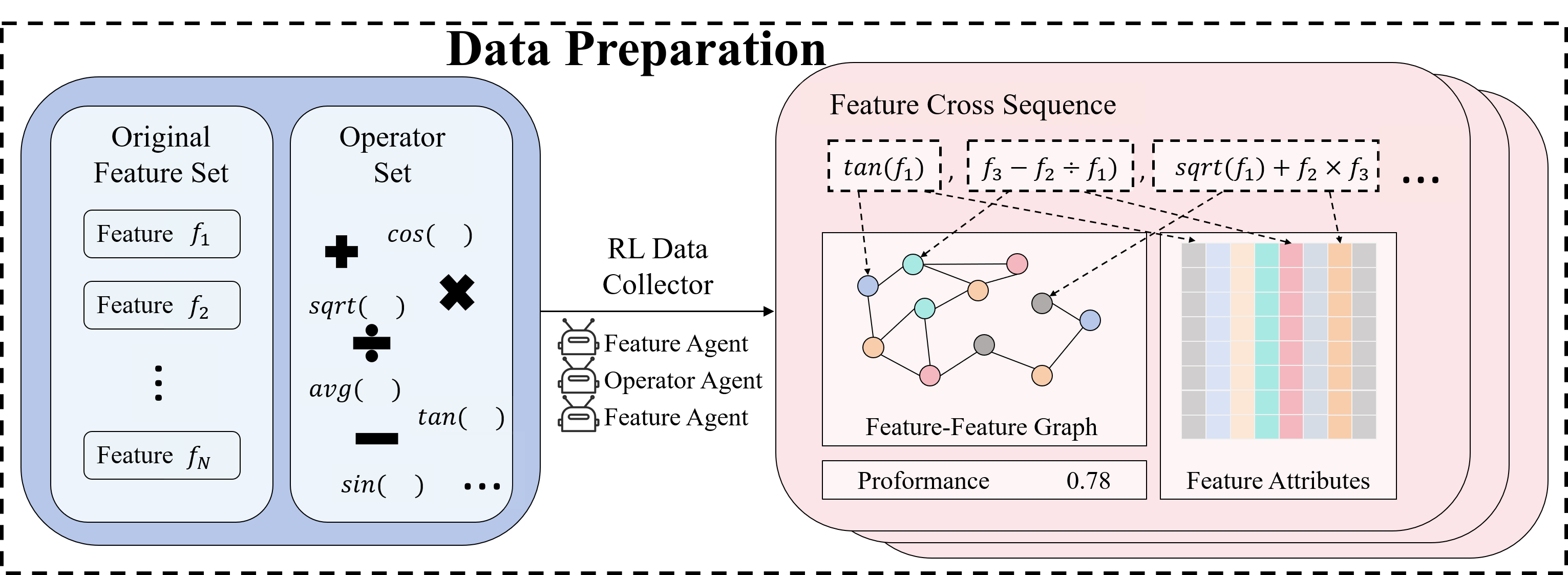}
\vspace{-0.5cm}
\caption{Data preparation pipeline. RL agents are employed to collect exemplary and diversified training data.}
\vspace{-0.5cm}
\label{data_prepare}
\end{figure}

\section{RL Agents for Data Collection}
Before modeling, we first construct a supervised knowledge base by collecting various transformed feature sets (specifically, different combinations of feature crosses) and their corresponding performances (e.g., accuracy) on a specific downstream task (e.g., regression or classification). This forms the observation values needed to build a continuous representation space. To construct this space, we can manually or randomly generate feature sets through cross-operations and evaluate their performance on a controlled task, which provides the training data for learning the feature set-to-vector encoder.
However, the three aspects of training data are critical:  1) \textit{large volume}:  sufficient training data;  2) \textit{high quality}: high-accuracy feature set cases as successful experiences;  3) \textit{high diversity}: training data should not ignore random and failure cases.

\noindent\textbf{Leveraging reinforcement learning to explore high quality and diverse training data.}
We leverage reinforcement learning (RL) to automatically generate diverse and high-performing feature sets as training data for our representation model (\textbf{Figure~\ref{data_prepare}}). Specifically, we design RL agents that iteratively perform feature crossing, guided by a reward signal based on downstream task performance. This process balances exploration and exploitation, enabling the collection of a large number of transformed feature sets paired with their performance metrics. These feature set–performance pairs form the supervised data used to train the feature set encoder. More details of the RL setup and training process are provided in \textbf{Appendix~\ref{appendix:RL}}.

\begin{figure}[h]
\centering
\includegraphics[width=\linewidth]{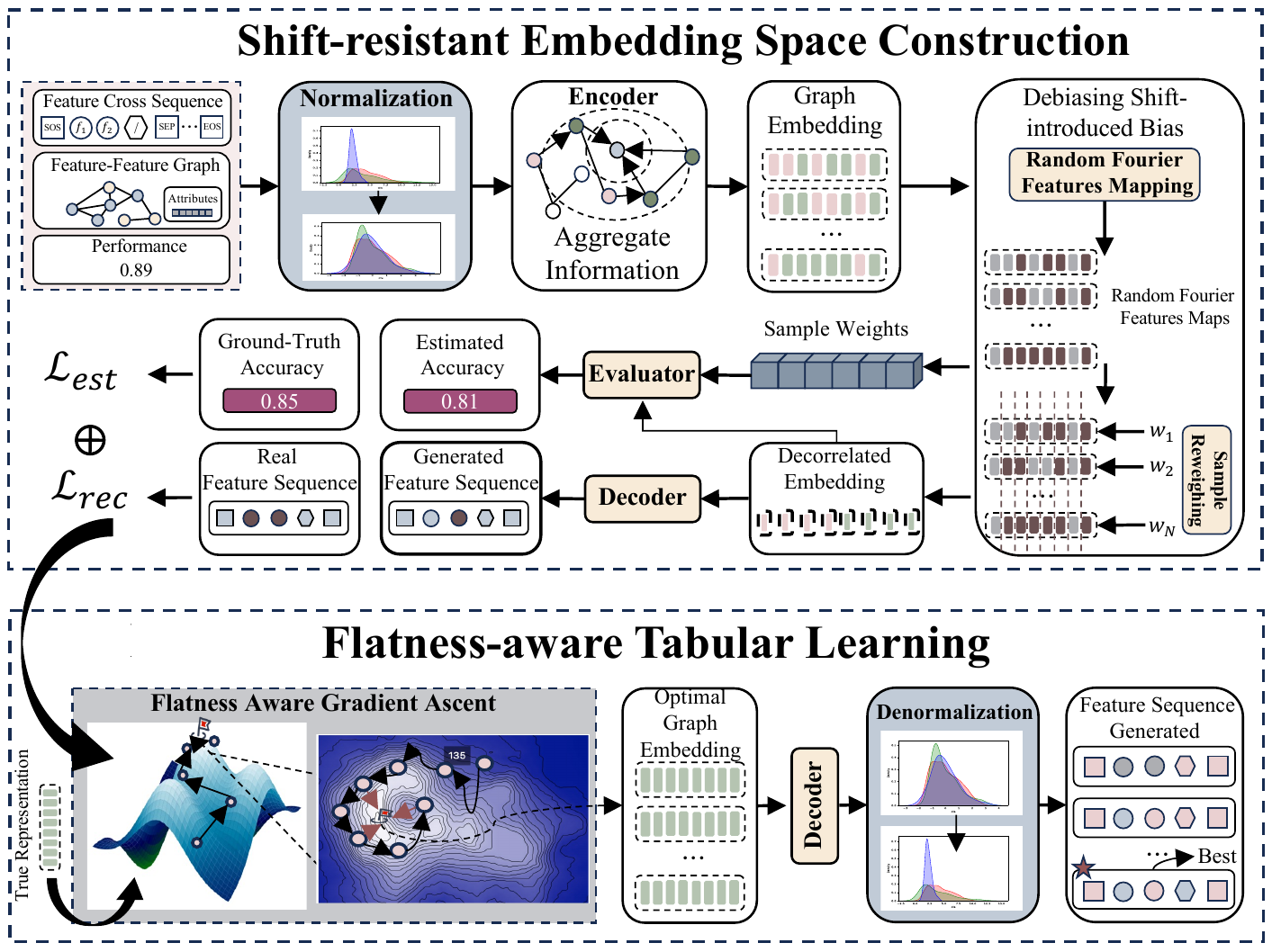}
\vspace{-0.5cm}
\caption{An overview of our proposed framework.}
\vspace{-0.2cm}
\label{framework}
\end{figure}

\section{The Representation-Generation Framework}
\subsection{Framework Overview}
\textbf{Figure~\ref{framework}} shows our Shift-Aware Feature Transformation framework (\textbf{SAFT}) for tabular learning, which includes three components: 
1) shift-resistant feature set representation;  
2) flatness-aware tabular learning;  
3) integrated pre-normalization and post-denormalization.  
To achieve a shift-resistant representation, we develop two insights: 1) feature sets as dynamic feature-feature interaction graphs and 2) a combination of sample reweighing and embedding dimension orthogonality for invariant representation. 
To achieve flatness-aware tabular learning, we incorporate flatness-aware gradient ascent, which seeks flatter optima in the embedding space for enhanced shift robustness. Finally, we integrate normalization in preprocessing and denormalization in postprocessing to further alleviate the impacts of shifts. We provide an algorithm in \textbf{Appendix~\ref{appendix:overall}} to clarify the overall process.

\subsection{Shift-resistant Feature Set Representation}
To find the best transformed feature set, our key insight is to effectively represent explored transformed feature sets and corresponding performances in an embedding space. Using training data composed of transformed feature sets and their predictive performances, we train an encoder and a decoder. These components map transformed feature sets to vectors and vice versa, creating a continuous embedding space. We believe that the best transformed feature set, termed feature knowledge, is described in this continuous embedding space, and thus can be efficiently identified by gradient optimization. As a result, we don't need to explore exponentially growing possibilities of feature combinations.
Although training-testing shifts of data points may distort the embedding space, it is learnable. Anti-shift mechanisms can rectify and adjust the distortion and ensure reliable and transferable feature knowledge in the testing set.

\subsubsection{The Encoder-Evaluator-Decoder Framework}

\noindent\blackcircnum{1} \ul{\textit{Encoder for Learning Dynamic Feature-Feature Interaction Graph Embedding.}}

Representing transformed feature sets is challenging due to strong inter-feature dependencies, which are often ignored by traditional methods. 
Simple aggregation of statistics or sequence-based encoding treats features as independent, leading to biased representations and suboptimal feature set selection.
We propose modeling a transformed feature set as a graph, where each node represents a feature, node attributes correspond to feature values, and edges reflect pairwise similarities (e.g., cosine similarity). 
This enables capturing feature-feature interactions directly. 
Although graph neural networks (e.g. GCNs) can embed such graphs, they assume fixed sizes and fixed topologies, making them unsuitable for feature sets with varying sizes and connections.
Inspired by~\cite{hamilton2017inductive}, we adapt node sampling and multi-hop neighborhood aggregation for dynamic feature-feature graph embedding.  
At each step, each node updates its representation by combining its own and neighbors' previous embeddings, followed by a nonlinear transformation via a fully connected layer.  
We normalize the output at each step and iterate to produce the final embedding.  
Formally, given a transformed feature set, we construct a feature-feature similarity graph $\mathcal{G}$ and encode it using an encoder $\phi_\theta$, yielding $E = \phi_\theta(\mathcal{G})$.

\noindent\blackcircnum{2} \ul{\textit{Evaluator for Estimating Feature Set Performance.}}
Our goal is to learn an embedding space that accurately represents transformed feature sets, allowing us to find the optimal embedding yielding higher performance, which is then used to reconstruct the optimal transformed feature set. 
To strengthen the expressiveness of the transformed feature set embedding, we incorporate a performance evaluator as a downstream task following the feature-feature graph encoder. 
Specifically, we employ a feedforward network to take an embedding as input and predict the performance of a transformed feature set, by minimizing the mean squared error between estimated performances and ground-truth. 
Formally, the evaluator loss is $\mathcal{L}_{est} = \frac{1}{N}\sum_{i=1}^N(p_i -\omega_{\theta}(E_i))^2$,  where $N$ is the total number of transformed feature sets in training data, $i$ indexes each transformed feature set, $p_i$ is the ground-truth performance corresponding to the transformed feature set, $\omega_{\theta}$ is the evaluator function, and $E_i$ is a specific graph embedding.

\noindent\blackcircnum{3} \ul{\textit{Decoder.}}
The primary objective of the decoder is to output a feature cross sequence (deemed as a token sequence) given a feature-feature graph embedding. 
The decoder provides two main functionalities: 1) Ensuring embedding faithfulness: the embedding of a transformed feature set should be able to fully reconstruct its corresponding feature cross sequences; 2) Generating the optimal transformed feature set:  a well-trained decoder can be used to generate the feature cross sequence of the optimal embedding.
Our decoder is structured by a single-layer LSTM and a softmax layer, where the LSTM recursively learns hidden states over the current-previous dependent feature cross sequence, and the softmax layer is used to estimate the token probability when generating each token. 
Specifically, given $\psi$ as a single-layer LSTM, the distribution of the single $i$-th token is 
$
    P_{\psi}(\gamma_i|E, \Upsilon_{<i}) = \frac{\rm{exp}(s_j)}{\sum_M\rm{exp}(s)}
$. 
where, $\Upsilon$ is a feature cross sequence with $M$ token length, $\gamma_i$ is the $i$-th token in $\Upsilon$, and $s_j$ is the $j-\rm{th}$ output of the softmax layer. 
Finally, we aim to minimize the negative log-likelihood of decoding a feature cross sequence as the reconstruction loss: $\mathcal{L}_{rec} =  -\sum_N\sum_{i=1}^M log P_{\psi}(\gamma_i|E, \Upsilon_{<i})$. 

\noindent\blackcircnum{4} \ul{\textit{The Joint Objective Function.}}
We optimize the joint loss of the encoder,  evaluator, and decoder. Our objective is: $\mathcal{L}_{est}+\gamma\mathcal{L}_{rec}$, where $\gamma$ is a hyperparameter that balances the loss. In this way, we enforce the encoder-evaluator-decoder structure to learn a faithful embedding that accurately predicts the performance of a transformed feature set and decode the corresponding feature cross sequence. 

\subsubsection{Correcting Training Biases under Distribution Shifts.}

\noindent\blackcircnum{1} \ul{\textit{Motivation: Toward Debiased Representations under Distribution Shifts.}} 
In open environments, distribution shifts introduce bias into the learned embedding space by entangling invariant (true) and spurious (false) representations. Their correlation can mislead optimization and reduce robustness.
A common mitigation is to decorrelate embeddings by minimizing the Frobenius norm of the covariance matrix, but this approach assumes structural orthogonality and may cause information loss.
We aim to suppress spurious bias by amplifying invariant signals. Inspired by stable sample weighting~\cite{shen2020stable} and robust reweighting~\cite{athey2018approximate}, we propose a bilevel framework: the inner loop learns sample weights to decorrelate embeddings, and the outer loop reweights losses for graph-based feature evaluation. This jointly mitigates bias and improves generalization under distribution shift.

\setlength{\textfloatsep}{2pt}
\begin{algorithm}[h]
    \caption{Shift-resistant Bilevel Training}
    \label{alg_1}
    \begin{algorithmic}[1]
    \REQUIRE EPOCH\_1 and EPOCH\_2 
    \ENSURE Learned Embedding Space
    \FOR {out\_loop\_epoch $\gets$ 1 to EPOCH\_1}
    \STATE Forward propagate

    \FOR {inner\_loop\_epoch $\gets$ 1 to EPOCH\_2}
    
    \STATE Optimize the sample weighting by Equation~(\ref{weights}) 
    \ENDFOR
    \STATE Back propagate with the weighted joint loss $\mathcal{L}= \mathcal{L}_{est} +  \gamma \mathcal{L}_{rec}$ to joint train 
    \ENDFOR
    \\
    \end{algorithmic}
\end{algorithm}
\noindent\blackcircnum{2} \ul{\textit{The Shift-resistant Bilevel Training.}} 
We introduce a two-loop algorithm to correct the training biases. \ul{1) For the inner-loop}, our objective is to learn the best sample weights to minimize the squared Frobenius norm as follows.
\begin{equation}
\label{weights}
\mathbf{R}^{\ast} = \mathop{\arg\min}_{\mathbf{R}} \sum_{i<j} \left\| \hat{\mathbf{C}}^{\mathbf{R}}_{E_{\ast i}, E_{\ast j}} \right\|^2_{F},  
\end{equation}
{\tiny
\begin{equation*}
\begin{aligned}
\text{\small where } \hat{\mathbf{C}}^{\mathbf{R}}_{E_{\ast i}, E_{\ast j}} = \frac{1}{N-1} \sum_{n=1}^{N} \bigg[
    \left(r_n f(E_{ni}) - \frac{1}{N} \sum_{m=1}^{N} r_m f(E_{mi})\right)^\top \\
    \cdot \left(r_n g(E_{nj}) - \frac{1}{N} \sum_{m=1}^{N} r_m g(E_{mj})\right)
\bigg].
\end{aligned}
\end{equation*}
}

Here, $\hat{\mathbf{C}}^{\mathbf{R}}_{E_{\ast i}, E_{\ast j}}$ is the partial cross-covariance matrix, $\mathbf{R} = \{r_n\}^N_{n=1}$ is the graph weight vector, $r_i$ is the weight of $i$-th feature subset graph and we constrain $\sum_{n=1}^N r_n= N$.
$f(\cdot)$ and $g(\cdot)$ are random Fourier feature functions, $E_{\ast i}$ and $E_{\ast j}$ denote different dimensions of the same training sample.
\ul{2) For the outer-loop training,} we use the updated weights of the training samples $\mathbf{R}^{\ast}= \{r_n^{\ast}\}^N_{n=1}$ to reweight the estimation loss of each transformed feature set, given by $\mathcal{L}_{est}=\sum_{i=1}^{N}r_i^{\ast}(p_i -\omega_{\theta}(E_i))^2$. 
Then, we combine the new estimator loss and the decoder loss $\mathcal{L}_{rec}$ to obtain the weighted joint loss: $ ~\mathcal{L}= \mathcal{L}_{est} +  \gamma \mathcal{L}_{rec}$. The \textbf{Algorithm~\ref{alg_1}} provides a detailed pseudocode.

\subsection{Flatness-aware Generation}

After learning the embedding space of transformed feature sets,  the embedding space accurately captures and represents all transformed feature sets, including the unobserved best transformed feature set. Hence, we utilize gradient ascent to identify the best embedding vector with the highest performance to decode the optimal transformed feature set. 

\noindent\textbf{Step 1: Gradient-ascent Optimization to Identify the Optimal Feature Set Embedding.}
Another benefit of learning an embedding performance evaluator in the representation step is that it enables differentiable gradient optimization. Specifically, we extract the gradient from the evaluator in the direction of maximizing the performance of the feature set. Formally, the gradient-ascent is defined by: $\hat{E} = E+\eta \frac{\partial \omega_{\theta}}{\partial E}$,  where $\omega_{\theta}$ is the evaluator,  $\mathbf{\hat{E}}$ is the optimal embedding, $\eta$ is the step size.  
In the experiments, we select top-T transformed feature set embeddings in training data as the initialization seeds for gradient ascent. 
Therefore, we can identify T-improved embedding vectors as the optimal embedding set $\hat{\mathcal{E}}= \{\hat{E}^t\}_{t=1}^{T}$.

\setlength{\textfloatsep}{2pt}
\begin{algorithm}[h]
    \caption{Flatness-Aware Gradient Ascent}
    \label{alg}
    \begin{algorithmic}[1]
    \REQUIRE Initialized embedding $E$, LR bounds $\eta_1$, $\eta_2$, \\
    cycle length $c$, number of iterations $n$
    \ENSURE Optimal embedding $\hat{E}$

    \STATE $\hat{E} \gets E$
    \FOR {$i$ = 1, 2, ..., n}
    \STATE $\eta \gets \eta(i)$~~ \{ Calculate LR for the iteration, $\eta_1 <\eta(i)<\eta_2$ \}
    \STATE $E\gets E+\eta \frac{\partial \omega_{\theta}}{\partial E}$ ~~\{ Stochastic gradient update \}
    \IF{mod ($i$, $c$) = 0}
    \STATE $n_{\rm{models}}\gets \frac{i}{c}$~~\{ Number of models \}
    \STATE $\hat{E} \gets \frac{\hat{E}\cdot n_{\rm{models}}+E}{n_{\rm{models}}+1}$ ~~\{ Update average \}
    \ENDIF

    \ENDFOR
    \\
    \end{algorithmic}
\end{algorithm}
\noindent\textbf{Step 2: Incorporating Flatness-awareness in Gradient Ascent to Mitigate Shifts.}
In gradient optimization, the flatness of the loss landscape has been shown to exhibit a close connection with distribution shift resistance both theoretically and empirically. 
When there exists a distribution shift between training and testing data, the embedding of the optimal transformed feature set in the test samples does not coincide with the optimal transformed feature set found in the training samples. 
Flatness ensures that the loss does not increase significantly in a neighborhood around the identified minimum. 
Therefore, flatness leads to distribution shift resistance because the loss in the test examples does not increase significantly. 
Inspired by~\cite{izmailov2018averaging, garipov2018loss}, we take advantage of loss flatness and develop a flatness-aware gradient ascent approach. 

Specifically, unlike the classic gradient ascent, we propose to enforce a search process that oscillates around the optimal embedding to collect more suboptimal embedding vectors. 
This suboptimal embedding set can pinpoint a flat region where the true optimal point is located for the test set.
We then aggregate all the suboptimal points as the final averaged embedding that represents the center of the flat region in the loss landscape.
However, averaging all the suboptimal points of each gradient ascent iteration introduces huge computational costs and may include the non-optimal points, which are far away from the optimal neighborhood. We therefore develop a cyclic scheme, in which a cycle includes multiple gradient ascent iterations, and we only average the suboptimal points at the end of a cycle. 
To avoid missing the optimal point when approaching the maximum, we utilize linearly decreasing learning rates within each cycle. 
In particular, \textbf{Algorithm \ref{alg}} shows that we linearly decrease the learning rate $\eta(i)$  from $\eta_1$ to $\eta_2$ over iterations during a cycle. In each iteration, we initialize the learning rate and update the embedding by gradient ascent. At the end of each cycle, we average the embedding (Line 7).

\noindent\textbf{Step 3: Decoding Embeddings to Reconstruct Optimal Feature Cross Sequences }
After obtaining the candidate embedding set $\hat{\mathcal{E}}$, we use the well-trained decoder $\psi$ to generate the feature cross sequences, i.e., $\hat{\mathcal{E}} \stackrel{\psi}{\rightarrow} \{\hat{\Upsilon}_i\}_{i=1}^T$. 
Specifically, the decoder iteratively generates the next token of a feature cross sequence, such as a feature token, an operator token, or a segmentation token ($<\text{sep}>$)  in an autoregressive manner until it produces an end-of-sequence token ($<\text{eos}>$). Finally, we divide the generated sequence into multiple segments using the ``$<\text{sep}>$" token and transform each segment into a feature, resulting in the optimal transformed feature set with the best estimated performance.

\setlength{\tabcolsep}{1.7mm}{
\begin{table*}[t]
\caption{Overall performance across various real-world datasets. The best and second-best outcomes are indicated by bold and underlined fonts, respectively. We measure the performance on classification (C) and regression (R) tasks using F1-score and (1-RAE) metrics, respectively. A higher value indicates higher quality of the feature transformation space.}
\vspace{-0.3cm}
\centering
\fontsize{7.8}{7.5}\selectfont
\begin{tabular}{c|ccc|cccccccccc|c}
\toprule \toprule
Dataset   & C/R & Samples & Features & RDG         & ERG         & LDA    & AFAT        & NFS         & TTG         & GRFG     & MOAT  &NEAT  &ELLM-FT  & \model \\
\midrule
Housing Boston  & R   & 506     & 13       & 0.375       & 0.366       & 0.146  & 0.387       & \underline{0.395} & 0.383      &0.361&\underline{0.395} & 0.367 & \underline{0.395} & \textbf{0.405}  \\
Airfoil     & R   & 1503    & 5        & 0.733       & 0.695       & 0.522  & \underline{0.742}     &\underline{0.742} & 0.738& 0.614 & 0.724 & 0.698 & \textbf{0.743} &\textbf{0.743} \\
openml\_586  & R   & 1000   & 25       &  0.542      &  0.536      & 0.104 & 0.540      & 0.543      & 0.543      & 0.334   & 0.616 & \underline{0.626} & 0.616  & \textbf{0.649}   \\
openml\_589  & R   & 1000   & 50       &  \underline{0.509}      &  0.472      &  0.099 &  0.467      & 0.470       &  0.469     &0.436   & 0.496 & 0.469 & \underline{0.509} & \textbf{0.582}   \\
openml\_607  & R   & 1000   & 50       &  0.306 &  0.310     &  0.054 &  0.344     &  0.358      & 0.354       &  0.371   & 0.424 & 0.358 & \underline{0.438} & \textbf{0.516}   \\
openml\_616  & R   & 500     & 50       & 0.197       & 0.311       & 0.014 & 0.342       & 0.344       & 0.343       & 0.459     & 0.369 & \underline{0.528} & \underline{0.528} &\textbf{0.534}    \\
openml\_618  & R   & 1000    & 50 & 0.421       & 0.391       & 0.058  & \underline{0.436} & 0.430        & 0.431       & 0.257         &0.427  & 0.391 & \textbf{0.468} & \textbf{0.468}   \\
openml\_620  & R   & 1000    & 25       &   0.510    &  0.464      & 0.025 & 0.475       & 0.464      & 0.462       & 0.495  &  \textbf{0.566} & \underline{0.545} & \underline{0.545} & \underline{0.545}   \\
openml\_637 & R   & 500     & 50       & 0.265       & 0.340       & 0.015 & 0.365       & 0.344       & 0.339       & 0.380    & 0.381 &0.380 & \underline{0.400} &\textbf{0.424}   \\
\midrule
Higgs Boston & C   & 50000   & 28       & 0.693       &  0.694 & 0.508  & 0.692       & 0.691       & 0.696       & 0.698  &  \underline{0.702} & 0.702 & \textbf{0.704} &  \textbf{0.704}  \\

SpectF  & C   & 267     & 44       & 0.674       & \underline{0.792}       & 0.651  & 0.643       & 0.716       & 0.672       & 0.728     & 0.766  & 0.776 & 0.792 & \textbf{0.799}   \\
UCI Credit & C   & 30000   & 25       & \underline{0.809} & 0.808       & 0.743  & 0.805       & 0.805       & 0.803       & 0.797          &0.808 & 0.809 & 0.803 &\textbf{0.816}     \\
Wine Quality Red& C   & 999     & 12       & 0.658       & 0.491       & 0.588  & 0.662       & 0.650        &0.675 & 0.668    &\underline{0.681}  & 0.622 & 0.680 &  \textbf{0.700}  \\
Wine Quality White & C   & 4900    & 12       & \underline{0.730}& 0.726       & 0.602  & 0.715 & 0.724      & 0.718& 0.680 &\underline{0.730} & 0.718 & 0.730 & \textbf{0.734}\\
PimaIndian & C   & 768    & 8       &0.679 &   0.761     &0.685   & 0.636 &  0.691     &0.734 & 0.689 & \underline{0.763} & 0.754 & 0.760 &\textbf{0.780} \\
Geman Credit & C   & 1000    & 24       & 0.697&   0.662     & 0.693  & 0.656 &   0.698   &0.683 & 0.647 & 0.707 & \underline{0.723} & 0.701 & \textbf{0.743}\\
\bottomrule \bottomrule
\end{tabular}

\label{main_exp}
\end{table*}}
\begin{figure*}[htp]
        \vspace{-0.3cm}
        \centering
        \subfigure[SpectF]{\label{exp:spectf}\includegraphics[width=0.16\textwidth]{{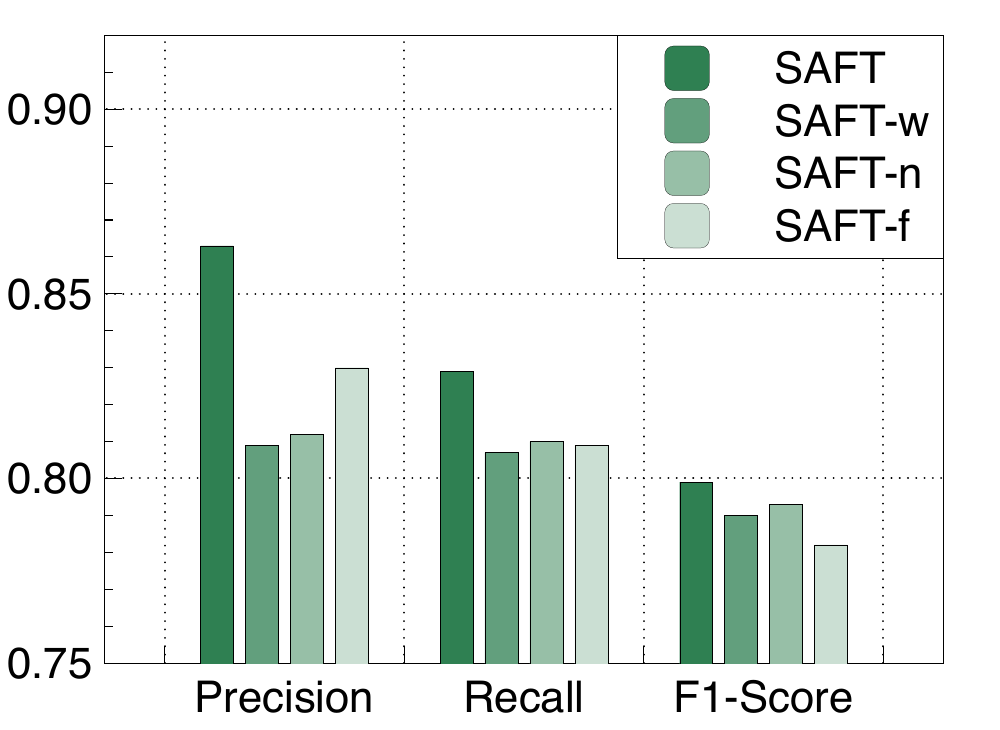}}}
        \subfigure[Wine White]{\label{exp:white}\includegraphics[width=0.16\textwidth]{{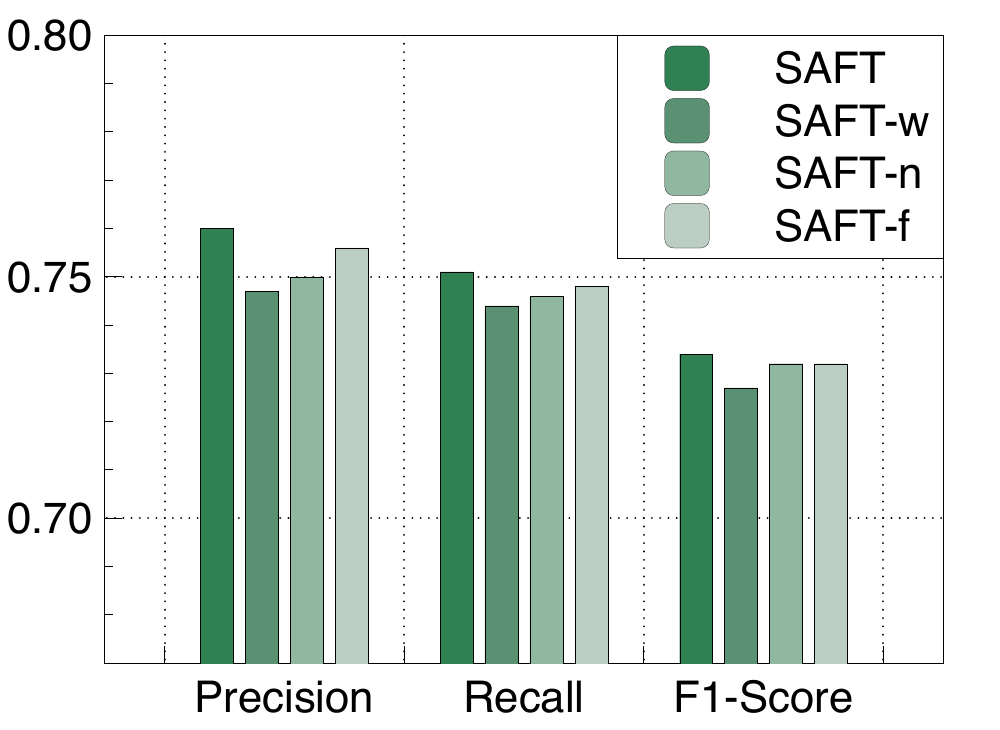}}}
        \subfigure[Wine Red]{\label{exp:red}\includegraphics[width=0.16\textwidth]{{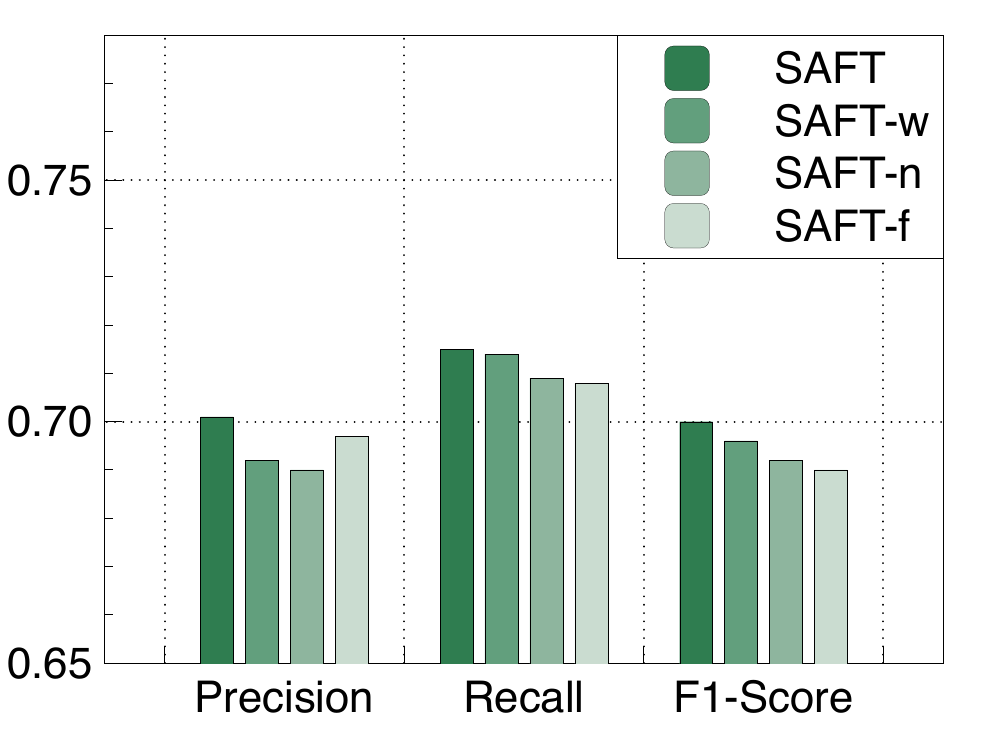}}}
        \subfigure[Openml\_616]{\label{exp:openml616}\includegraphics[width=0.16\textwidth]{{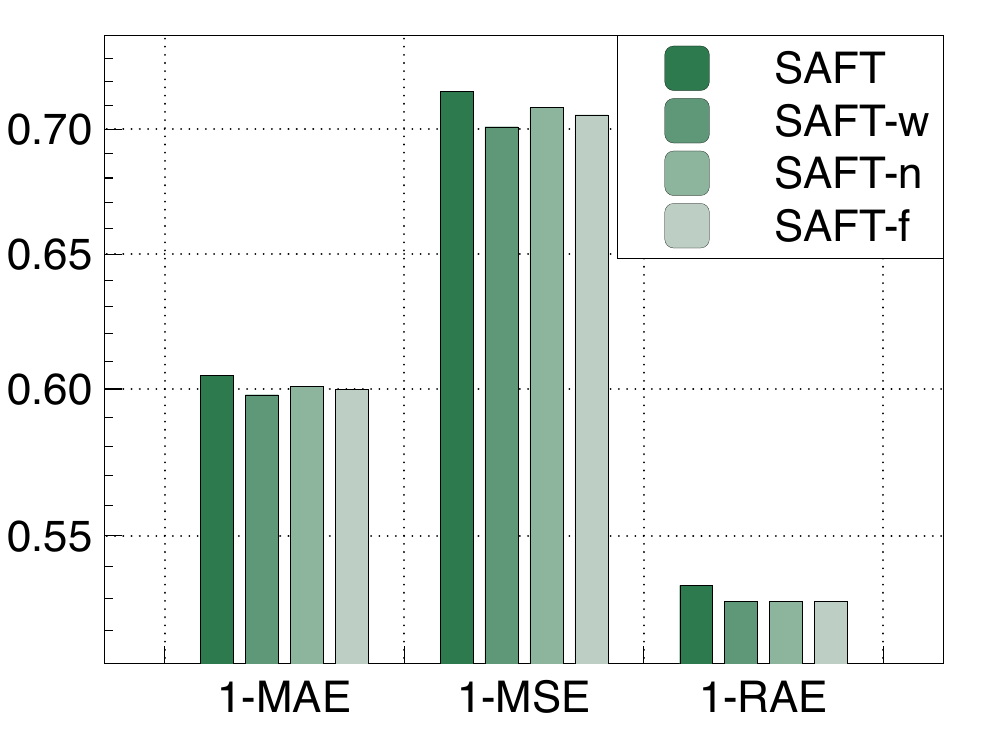}}}
        \subfigure[Openml\_618]{\label{exp:openml618}\includegraphics[width=0.16\textwidth]{{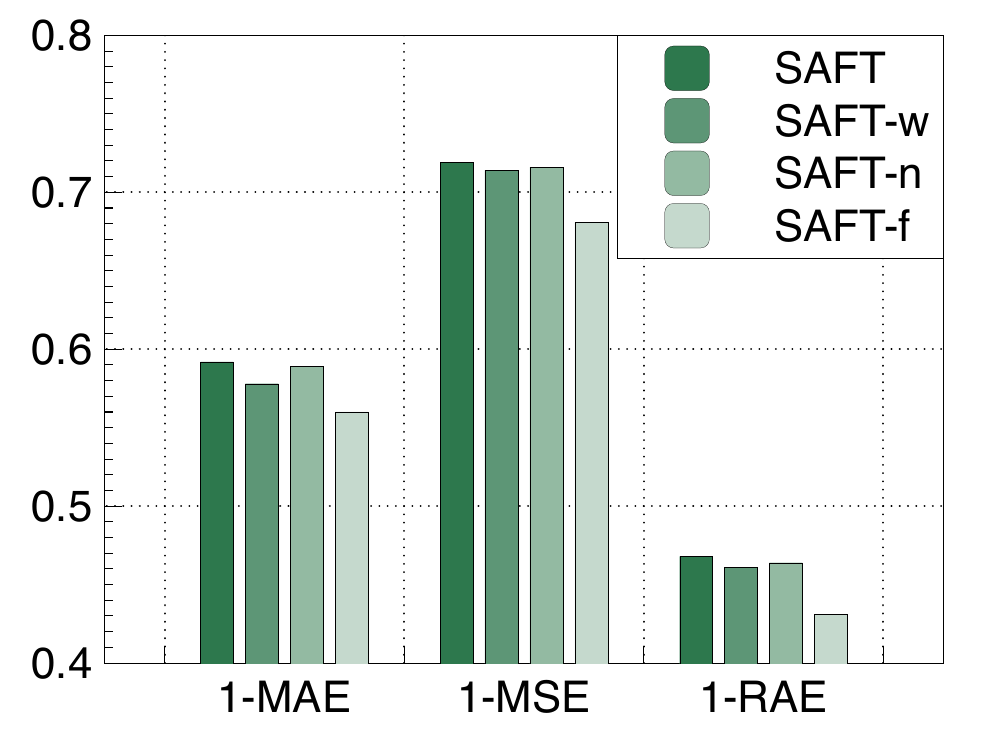}}}
        \subfigure[Openml\_637]{\label{exp:openml637}\includegraphics[width=0.16\textwidth]{{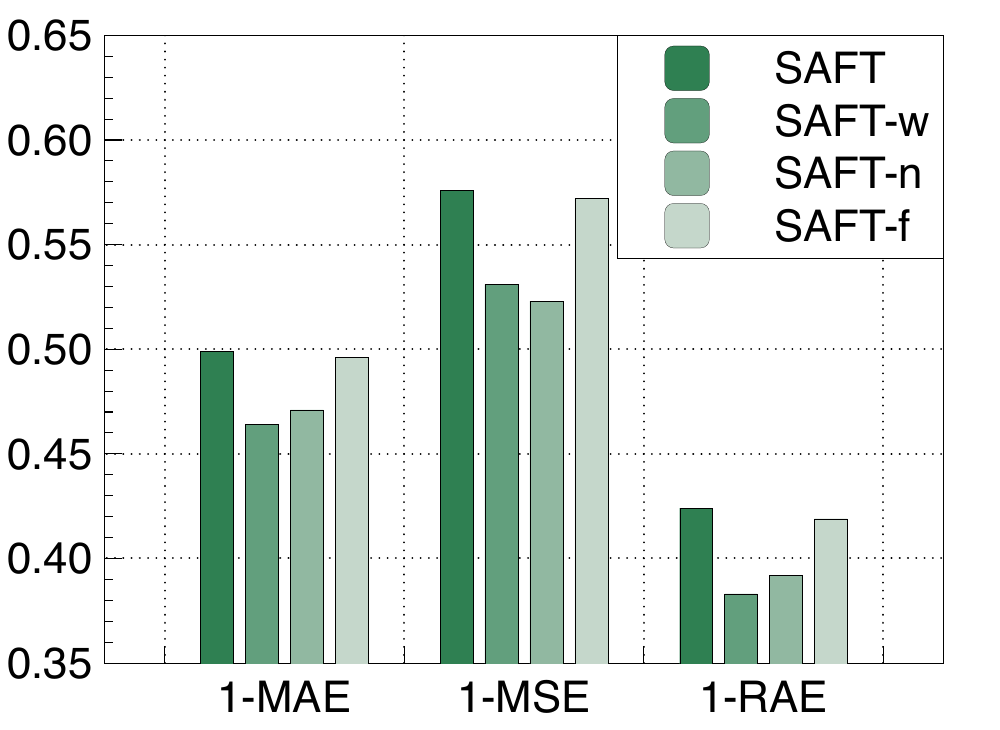}}}
        \vspace{-0.3cm}
        \caption{Results on the impact of normalization, Flatness Aware Gradient Ascent, and reweighting.}
        \vspace{-0.4cm}
        \label{Alabtion_exp}
\end{figure*}

\subsection{Integrating Normalization-Denormalization into Pre- and Post-Processing}
In addition to embedding dimension orthogonality and flatness-aware optimization, we propose a data-centric perspective to alleviate distribution shift by aligning and recovering the distribution gaps between training and test data in pre- and post-processing.
We incorporate a normalization and denormalization mechanism. Specifically, we first normalize (e.g., z-score) both training and testing data before feature transformation, obtaining normalized data with zero mean and unit variance.
We train our method on normalized training data to generate the optimal feature cross-sequence under a normalized distribution.
Next, we apply this learned sequence to transform the normalized test data.
Then, we denormalize (e.g., reverse z-score) the transformed test data to restore the original distribution.
This approach enables feature transformation learning in a normalized distribution and avoids shift-induced bias.

\section{Experimental Results}
\subsection{Experimental Setup}
\noindent\textbf{Datasets.}
We perform experiments on 16 publicly available datasets from UCI~\cite{Dua:2019} and OpenML~\cite{OpenML2013}, which are widely used to evaluate tabular learning methods. The statistics of all datasets are described in \textbf{Table~\ref{main_exp}}.
To evaluate \model's robustness to distribution shifts, we iteratively split the dataset (80\%/20\%) along each feature and apply the Kolmogorov–Smirnov test~\cite{berger2014kolmogorov} at 95\% confidence. If no shift is detected in any feature, the dataset is excluded.

\noindent\textbf{Evaluation Metrics.}
To minimize the influence of downstream models, we employ Random Forests (RF) for all tasks. Evaluation uses F1, Precision, and Recall for classification, and 1-RAE, 1-MAE, and 1-MSE~\cite{hill2012encyclopedia,hodson2022root} for regression, where higher scores reflect better feature transformations.

\noindent\textbf{Baselines.}
We compare our framework with 10 widely used feature transformation methods:
1) \textbf{Random Generation (RDG)};
2) \textbf{Essential Random Generation (ERG)};
3) \textbf{Latent Dirichlet Allocation (LDA)}~\cite{blei2003latent};
4) \textbf{AutoFeat Automated Transformation (AFAT)}~\cite{horn2020autofeat};
5) \textbf{Neural Feature Search (NFS)}~\cite{chen2019neural};
6) \textbf{Traversal Transformation Graph (TTG)}~\cite{khurana2018feature}
7) \textbf{Group-wise Reinforcement Feature Generation (GRFG)}~\cite{wang2022group};
8) \textbf{reinforceMent-enhanced autOregressive feAture Transformation (MOAT)}~\cite{wang2023reinforcement};
9) \textbf{uNsupervised gEnerative feAture Transformation (NEAT)}~\cite{ying2024unsupervised}
10). \textbf{Evolutionary Large Language Model framework for automated Feature Transformation (ELLM-FT)}~\cite{gong2025evolutionary}
More details about the baselines are provided in \textbf{Appendix~\ref{appendix:exp_setup}}.

\noindent To make the experiment reproducible, we provide the details about the parameters and environment in \textbf{Appendix~\ref{appendix:exp_setup}}.

\subsection{Performance Results}
In this experiment, we compare the feature transformation performance of \model\ with baseline models.
\textbf{Table~\ref{main_exp}} shows the overall comparison results in terms of F1 score and 1-RAE.
We observe that \model\ consistently outperforms baseline models across all datasets.
The underlying driver for this observation is that \model\ can compress the feature learning knowledge into a robust embedding space through shift-resistant embedding learning, and smoothly search for the optimal feature space via flatness-aware weight averaging.
This helps preserve generalizable patterns across shifts.
Moreover, we notice that the second-best model varies among different cases.
A potential reason for this observation is that baseline approaches overlook the impact of distribution shifts, resulting in an unstable identification of the optimal feature space.
This sensitivity reflects a lack of robustness under varying distributions.
This experiment underscores that \model\ effectively captures invariant knowledge in feature learning against distribution shifts and produces a robust transformed feature space.

\vspace{-0.2cm}
\subsection{Ablation Study}
To evaluate the necessity of various components in \model, we develop three model variants: 1) \model-f, which excludes the Flatness-Aware Gradient Ascent and does not perform weight averaging; 2) \model-n, which omits the normalization process; 3) \model-w, which removes the optimization of graph weights. 
We select three classification and three regression datasets for conducting experiments. \textbf{Figure~\ref{Alabtion_exp}} shows comparison results. 
First, we observe that, in all situations, the performance of the three model variants declines compared to \model.
A potential reason is that disregarding distribution-shift-aware strategies fails to consider the variation between training and testing sets, resulting in suboptimal feature transformation performance.
Moreover, \model\ can perform better compared to \model-w. 
This observation indicates that the learned graph weights help eliminate spurious correlations among features, thus enhancing the transformed feature space.
Furthermore, we find that \model outperforms \model-f. 
The underlying driver is that Flatness-Aware Gradient Ascent prevents the gradient search process from converging to locally optimal points, thereby smoothing and making the search process more robust.
Furthermore, \model-n is inferior to \model. This observation reflects that aligning statistical properties through normalization is an effective strategy to tackle the OOD issue in feature space.
Thus, these experiments demonstrate that each technical component of \model\ is indispensable for mitigating the distribution shift on feature transformation.

\begin{figure}[t]
\vspace{-0.2cm}
        \centering
        \subfigure[Precision]{\includegraphics[width=0.15\textwidth]{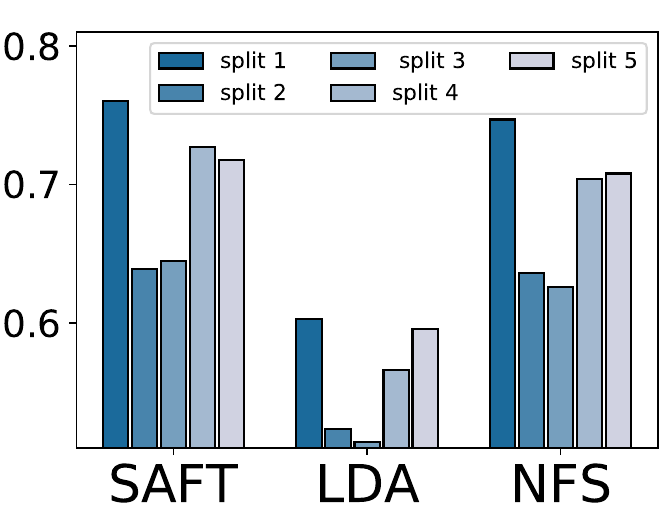}}
        \subfigure[Recall]{\includegraphics[width=0.15\textwidth]{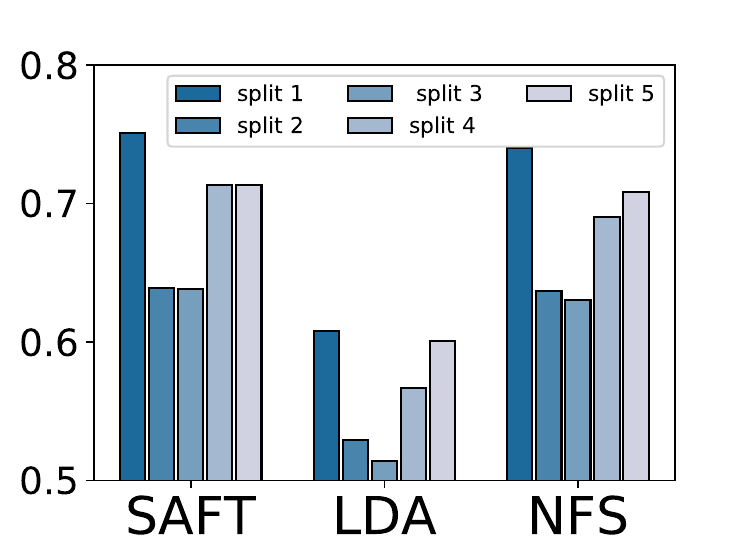}}
        \subfigure[F1 Score]{\includegraphics[width=0.15\textwidth]{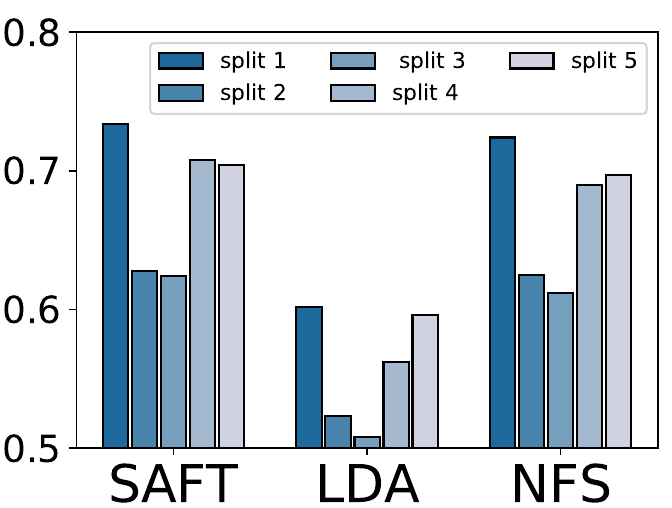}}
        \vspace{-0.3cm}
        \caption{Robustness of the dataset split methods.}
        \label{ro_1_exp}
\end{figure}
\subsection{Robustness Analysis}
\label{Robustness Analysis}
\noindent\textbf{Dataset Split Robustness.}
To check the robustness of \model\ for different data splits, we use five distinct splits on the Wine Quality White dataset for evaluation. Each split defines a unique 20\% proportion of the dataset as the testing set, moving sequentially through the dataset from beginning to end, with the remainder serving as the training set.
\textbf{Figure~\ref{ro_1_exp}} presents the comparison results in terms of F1-score, Precision, and Recall. We notice that despite potential distribution shifts introduced by different dataset split approaches, \model\ still demonstrates robust performance in addressing these shifts. 
The main reason for this observation is that 
\model\ can preserve the invariant and robust feature learning knowledge within the embedding space, enabling a smooth and stable search.
Such a learning mechanism can alleviate the impact of distribution shifts across different data splits.
Hence, this experiment shows that \model\ exhibits robust performance in addressing distribution shift, regardless of the dataset split method.

\setlength{\tabcolsep}{2.3mm}{
\begin{table}[tb]
\centering
\fontsize{7.8}{7.5}\selectfont
\caption{Robustness check on ML models for the Wine Quality White dataset.}
\vspace{-0.3cm}
\begin{tabular}{@{}c|cccccc@{}}
\toprule
        & RF             & SVM            & KNN            & DT             & LASSO          & Ridge          \\
        \midrule
RDG     & 0.658          & 0.632          & \textbf{0.629} & 0.630          & \underline{0.634}   & 0.606          \\
ERG     & 0.491          & 0.354          & 0.358          & 0.413          & 0.470          & 0.522          \\
LDA     & 0.588          & 0.439          & 0.512          & 0.579          & 0.390          & 0.410          \\
AFAT    & 0.662          & 0.582          & 0.443          & 0.663   & 0.596          & 0.587          \\
NFS     & 0.650          & 0.634          & 0.542          & 0.639          & \underline{0.634}    & 0.601          \\
TTG     & \underline{0.675}  & \underline{0.644}    & 0.525          & 0.602          & 0.630          & 0.620   \\
GRFG    & 0.668          & 0.313          & 0.502          & 0.616          & 0.551          & 0.540          \\
MOAT    & 0.588          & 0.359          & 0.512          & 0.629          & 0.590          & 0.606          \\
NEAT    & 0.653          & 0.630          & 0.500          & 0.633          & 0.630          & 0.601          \\
ELLM-FT & 0.662          & 0.634          & 0.542          & \underline{0.668}          & \textbf{0.659}          & \underline{0.638}         \\
\midrule
\model & \textbf{0.700} & \textbf{0.655} & \underline{ 0.556}    & \textbf{0.677} & \textbf{0.659} & \textbf{0.658} \\
\bottomrule
\end{tabular}
\label{exp:down}
\end{table}}

\noindent\textbf{Downstream Performance Robustness.}
To check the robustness of \model\ for distinct downstream ML models, we substitute the downstream model with Random Forest (RF), Support Vector Machine (SVM), K-Nearest Neighborhood (KNN), Ridge, LASSO, and Decision Tree (DT). 
\textbf{Table~\ref{exp:down}} shows the comparison results on the Wine Quality Red dataset in terms of the F1-score. 
We find that \model\ outperforms baselines in most cases. 
A potential reason for this observation is that  \model\ captures invariant and generalizable feature knowledge, leading to its superior generalization across various downstream ML models.
Thus, this experiment demonstrates the robustness of \model\ across various ML models.

\begin{figure}[h]
    \vspace{-0.3cm}
    \centering
    \subfigure[Wine White]{\label{training}\includegraphics[width=0.23\textwidth]{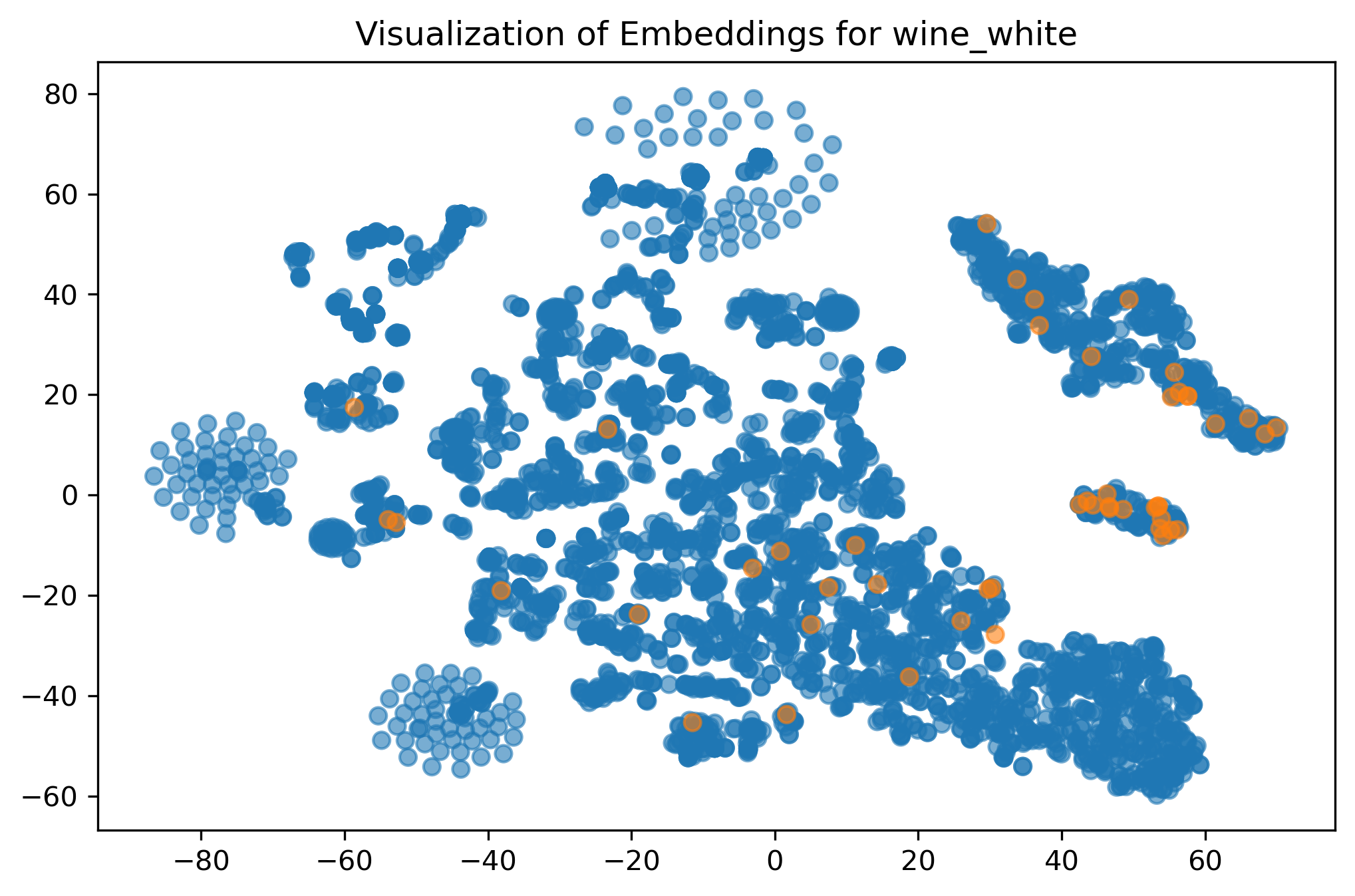}}
    \subfigure[UCI Credit]{\label{test}\includegraphics[width=0.23\textwidth]{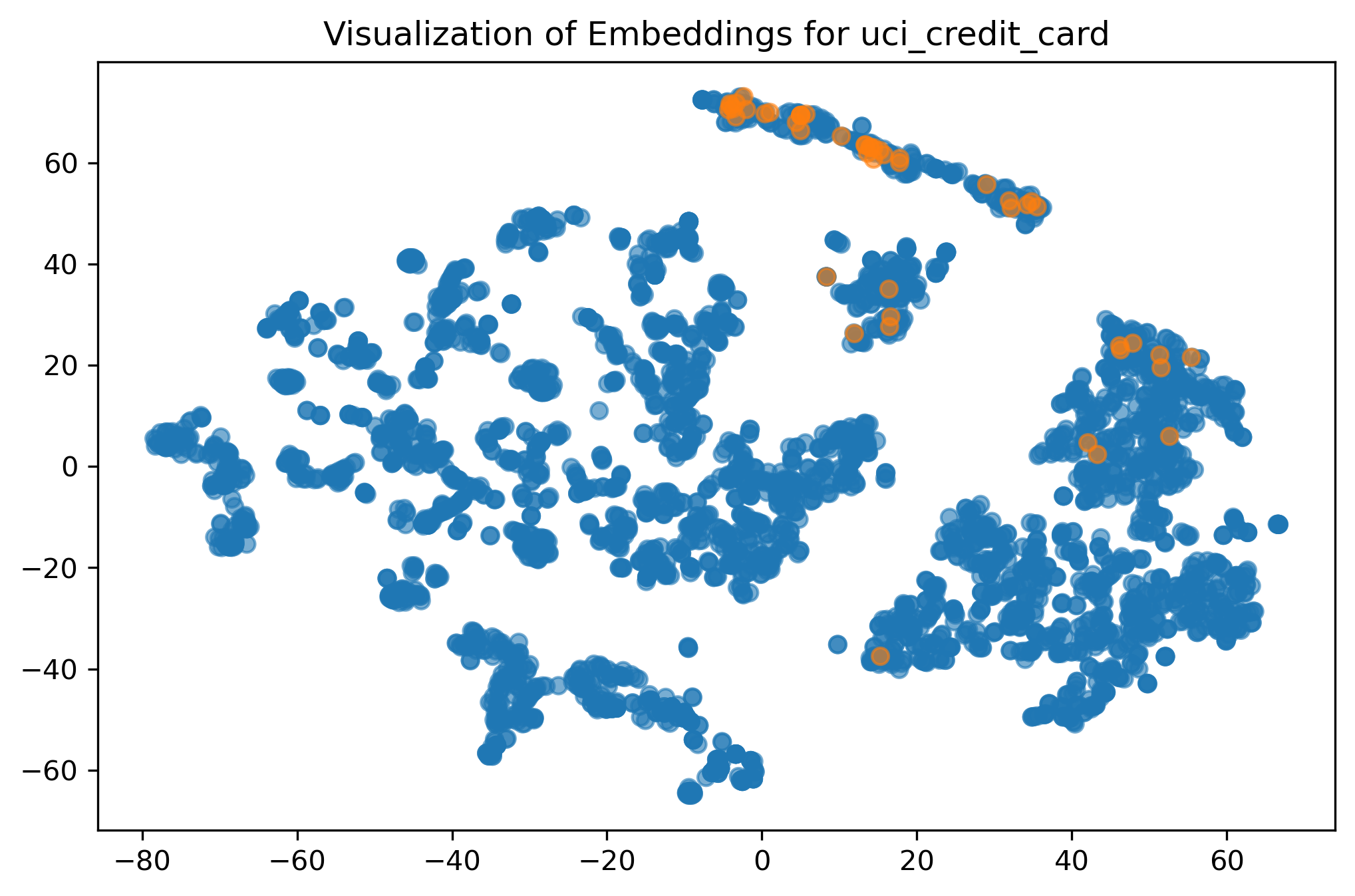}}
    \subfigure[Openml\_616]{\label{training}\includegraphics[width=0.23\textwidth]{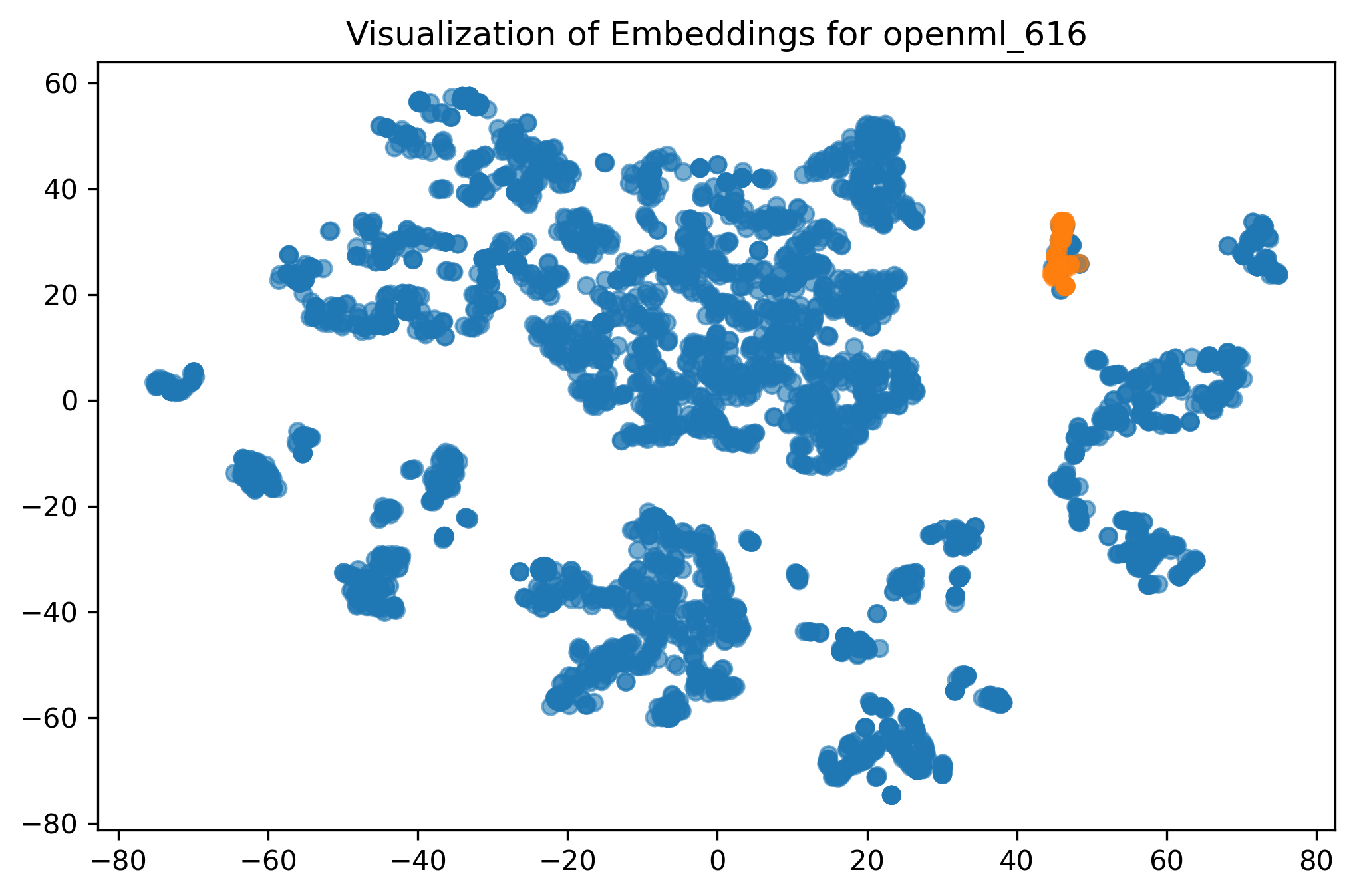}}
    \subfigure[Openml\_637]{\label{test}\includegraphics[width=0.23\textwidth]{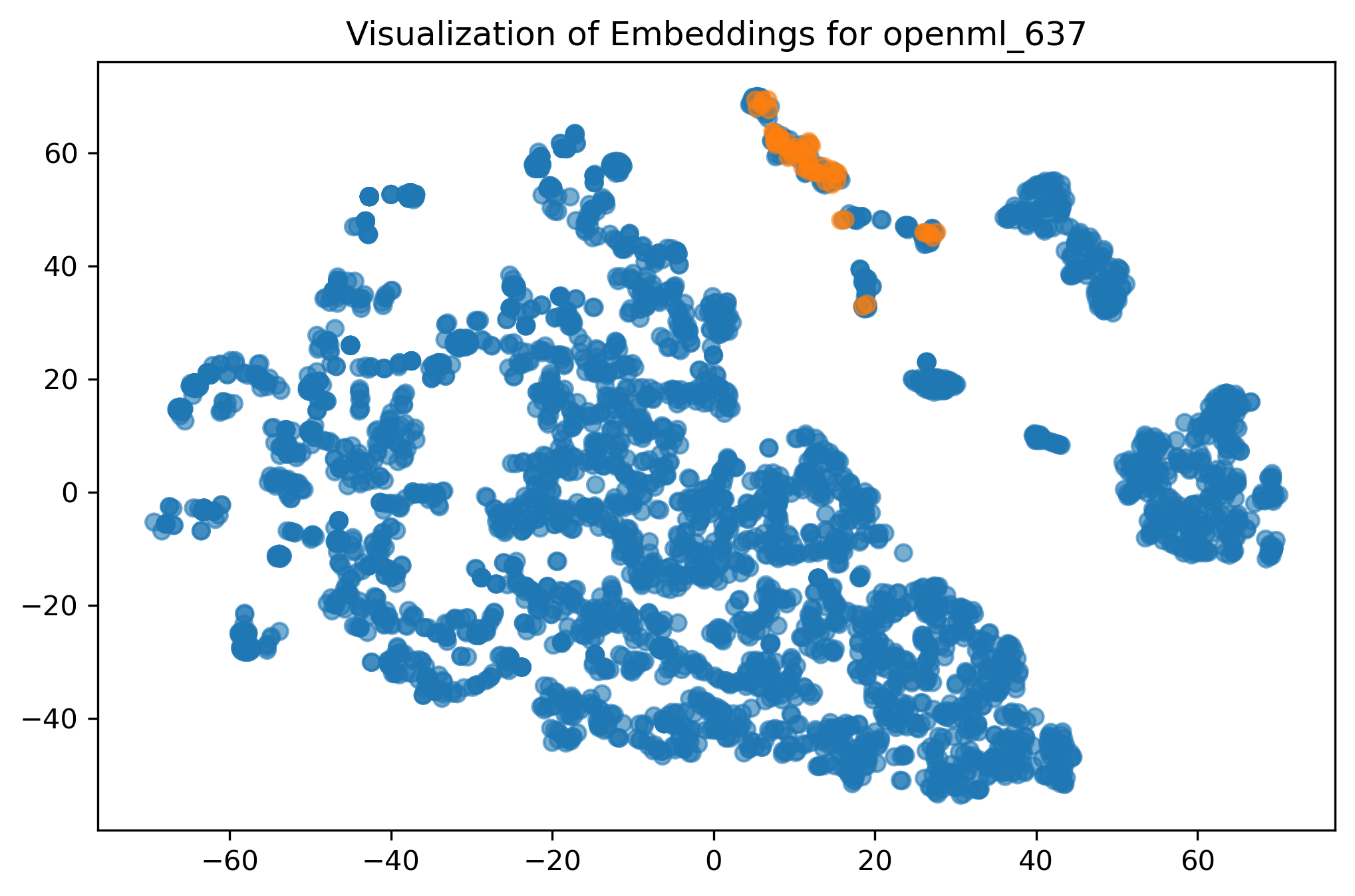}}
    \vspace{-0.3cm}
    \caption{Visualization of the converged embedding space. Each point represents a feature transformation sequence, with orange points indicating the top 50 embeddings ranked by downstream performance.}
    \vspace{-0.6cm}
    \label{space_visual}
\end{figure}

\subsection{Visualization for the Embedding Space}
In this section, we visualize the embedding space to validate the effectiveness of the intermediate encoding.
We collect latent embeddings from transformation records and apply t-SNE to project them into two dimensions.
\textbf{Figure~\ref{space_visual}} presents results on four datasets, where each point represents a transformation operation sequence.
Despite varying sequence lengths leading to spread-out distributions, we observe clear clustering patterns, particularly among top-performing sequences (orange points), which are densely grouped.
This suggests that high-quality transformations tend to concentrate in specific regions, enabling efficient gradient-based search when using these clusters as initialization.
These results demonstrate that the embedding space captures meaningful structure, supporting effective reconstruction and performance estimation.

To make the experiments more convincing, we conduct qualitative analysis, complexity comparisons with baselines, and complexity analysis of \model as presented in \textbf{Appendix~\ref{appendix:qualitative}, ~\ref{appendix:complexity_base}, and ~\ref{appendix:complexity}}, respectively.

\section{Related Work}
Feature Transformation aims to construct an enhanced feature space by transforming original features. Previous research can be categorized into three groups: 1) \textit{Expansion-reduction based approaches}~\cite{kanter2015deep,horn2020autofeat,khurana2016cognito}, where the original feature space is expanded through explicit or greedy mathematical transformations and subsequently reduced by selecting valuable features.
2) \textit{Evolution-evaluation approaches}~\cite{hu2024reinforcement,khurana2018feature,tran2016genetic}, where evolutionary algorithms or reinforcement learning models are employed to optimize the process of iteratively creating effective features and retaining significant ones.
3) \textit{Generative AI-based}~\cite{gong2025agentic,gong2025sculpting,gong2025unsupervised,gong2025evolutionary}, where the most appropriate model architecture is automatically identified to formulate AFT as an AutoML task. However, these methods face two challenges: 1) Generating high-order feature transformations is difficult; 2) Their transformation performance is often unstable. 
Distribution Shift Problems: arising from inconsistent distributions between the training set and test set, are prevalent in the real-world domains. Existing works to solve this problem mainly focus on time series analysis~\cite{fan2023dish}, computer vision~\cite{yu2023distribution}, and natural language processing~\cite{dou2022decorrelate}, etc. However, distribution shift in the context of feature transformation remains underexplored, despite its significant impact on the effectiveness of transformed features. To address this gap, we integrate three shift-aware mechanisms at different levels of our framework:
1) \textit{Shift-resistant representation}, which captures invariant representations under DSTL; 2) \textit{Flatness-aware generation}, which identifies robust embeddings during the search process; and 3) \textit{Shift-aligned pre- and post-processing}, which reduces distribution mismatches during data transformation. This work advances feature engineering~\cite{ying2024feature,gong2025neuro,ying2024revolutionizing,wang2024knockoff,bai2025privacy,wang2025llm}. Such progress benefits critical domains, from renewable energy forecasting~\cite{huo2025ct} to image translation~\cite{zhu2025image} and fraud detection~\cite{huo2025enhancing}, where reliable representations enhance sustainability, creativity, and financial security.

\section{Conclusion}
We propose the Shift-Aware Feature Transformation (\model) framework, which integrates representation learning, generation, and anti-shift mechanisms to enable robust feature transformation. Our contributions are summarized as follows:
First, we reformulate DSTL as a gradient-based optimization problem and design a representation-generation framework that enables the deployment of an anti-shift mechanism to solve the OOD problem.
Second, to address the distribution shift, we incorporate three mechanisms at different levels of the framework:
1) \textit{Shift-resistant representation}, to learn invariant features under DSTL;
2) \textit{Flatness-aware generation}, to improve embedding robustness during optimization;
3) \textit{Distribution-aligned pre- and post-processing}, to reduce train-test discrepancies.
Finally, empirical results demonstrate that \model\ generates a robust feature space and discovers an optimal transformed feature set using only training data, without requiring knowledge of the test distribution. This highlights its strong generalization ability and potential impact across open-ended domains.


\bibliography{aaai2026}

\begin{thebibliography}{48}
\providecommand{\natexlab}[1]{#1}

\bibitem[{Athey, Imbens, and Wager(2018)}]{athey2018approximate}
Athey, S.; Imbens, G.~W.; and Wager, S. 2018.
\newblock Approximate residual balancing: debiased inference of average treatment effects in high dimensions.
\newblock \emph{Journal of the Royal Statistical Society Series B: Statistical Methodology}, 80(4): 597--623.

\bibitem[{Bai et~al.(2025)Bai, Ying, Gong, Wang, and Fu}]{bai2025privacy}
Bai, H.; Ying, W.; Gong, N.; Wang, X.; and Fu, Y. 2025.
\newblock Privacy-preserving data reprogramming.
\newblock \emph{npj Artificial Intelligence}, 1(1): 10.

\bibitem[{Berger and Zhou(2014)}]{berger2014kolmogorov}
Berger, V.~W.; and Zhou, Y. 2014.
\newblock Kolmogorov--smirnov test: Overview.
\newblock \emph{Wiley statsref: Statistics reference online}.

\bibitem[{Blei, Ng, and Jordan(2003)}]{blei2003latent}
Blei, D.~M.; Ng, A.~Y.; and Jordan, M.~I. 2003.
\newblock Latent dirichlet allocation.
\newblock \emph{Journal of machine Learning research}, 3(Jan): 993--1022.

\bibitem[{Chen et~al.(2019)Chen, Lin, Luo, Li, Zhang, Xu, Dang, Sui, Zhang, Qiao et~al.}]{chen2019neural}
Chen, X.; Lin, Q.; Luo, C.; Li, X.; Zhang, H.; Xu, Y.; Dang, Y.; Sui, K.; Zhang, X.; Qiao, B.; et~al. 2019.
\newblock Neural feature search: A neural architecture for automated feature engineering.
\newblock In \emph{2019 IEEE International Conference on Data Mining (ICDM)}, 71--80. IEEE.

\bibitem[{Dou et~al.(2022)Dou, Zheng, Wu, Gao, Shan, Zhang, Wu, and Huang}]{dou2022decorrelate}
Dou, S.; Zheng, R.; Wu, T.; Gao, S.; Shan, J.; Zhang, Q.; Wu, Y.; and Huang, X. 2022.
\newblock Decorrelate irrelevant, purify relevant: Overcome textual spurious correlations from a feature perspective.
\newblock \emph{arXiv preprint arXiv:2202.08048}.

\bibitem[{Dua and Graff(2017)}]{Dua:2019}
Dua, D.; and Graff, C. 2017.
\newblock UCI Machine Learning Repository.

\bibitem[{Fan et~al.(2023)Fan, Wang, Wang, Wang, Zhou, and Fu}]{fan2023dish}
Fan, W.; Wang, P.; Wang, D.; Wang, D.; Zhou, Y.; and Fu, Y. 2023.
\newblock Dish-TS: a general paradigm for alleviating distribution shift in time series forecasting.
\newblock In \emph{Proceedings of the AAAI Conference on Artificial Intelligence}, volume~37, 7522--7529.

\bibitem[{Fan et~al.(2024)Fan, Zheng, Wang, Xie, Bian, and Fu}]{fan2024addressing}
Fan, W.; Zheng, S.; Wang, P.; Xie, R.; Bian, J.; and Fu, Y. 2024.
\newblock Addressing Distribution Shift in Time Series Forecasting with Instance Normalization Flows.
\newblock \emph{arXiv preprint arXiv:2401.16777}.

\bibitem[{Garipov et~al.(2018)Garipov, Izmailov, Podoprikhin, Vetrov, and Wilson}]{garipov2018loss}
Garipov, T.; Izmailov, P.; Podoprikhin, D.; Vetrov, D.~P.; and Wilson, A.~G. 2018.
\newblock Loss surfaces, mode connectivity, and fast ensembling of dnns.
\newblock \emph{Advances in neural information processing systems}, 31.

\bibitem[{Gong et~al.(2025{\natexlab{a}})Gong, Dong, Bai, Wang, Ying, and Fu}]{gong2025agentic}
Gong, N.; Dong, S.; Bai, H.; Wang, X.; Ying, W.; and Fu, Y. 2025{\natexlab{a}}.
\newblock Agentic Feature Augmentation: Unifying Selection and Generation with Teaming, Planning, and Memories.
\newblock \emph{arXiv preprint arXiv:2505.15076}.

\bibitem[{Gong et~al.(2025{\natexlab{b}})Gong, Li, Dong, Bai, Ying, Wang, and Fu}]{gong2025sculpting}
Gong, N.; Li, Z.; Dong, S.; Bai, H.; Ying, W.; Wang, X.; and Fu, Y. 2025{\natexlab{b}}.
\newblock Sculpting Features from Noise: Reward-Guided Hierarchical Diffusion for Task-Optimal Feature Transformation.
\newblock \emph{arXiv preprint arXiv:2505.15152}.

\bibitem[{Gong et~al.(2025{\natexlab{c}})Gong, Reddy, Ying, Chen, and Fu}]{gong2025evolutionary}
Gong, N.; Reddy, C.~K.; Ying, W.; Chen, H.; and Fu, Y. 2025{\natexlab{c}}.
\newblock Evolutionary large language model for automated feature transformation.
\newblock In \emph{Proceedings of the AAAI Conference on Artificial Intelligence}, volume~39, 16844--16852.

\bibitem[{Gong et~al.(2025{\natexlab{d}})Gong, Wang, Ying, Bai, Dong, Chen, and Fu}]{gong2025unsupervised}
Gong, N.; Wang, X.; Ying, W.; Bai, H.; Dong, S.; Chen, H.; and Fu, Y. 2025{\natexlab{d}}.
\newblock Unsupervised Feature Transformation via In-context Generation, Generator-critic LLM Agents, and Duet-play Teaming.
\newblock \emph{arXiv preprint arXiv:2504.21304}.

\bibitem[{Gong et~al.(2025{\natexlab{e}})Gong, Ying, Wang, and Fu}]{gong2025neuro}
Gong, N.; Ying, W.; Wang, D.; and Fu, Y. 2025{\natexlab{e}}.
\newblock Neuro-symbolic embedding for short and effective feature selection via autoregressive generation.
\newblock \emph{ACM Transactions on Intelligent Systems and Technology}, 16(2): 1--21.

\bibitem[{Hamilton, Ying, and Leskovec(2017)}]{hamilton2017inductive}
Hamilton, W.; Ying, Z.; and Leskovec, J. 2017.
\newblock Inductive representation learning on large graphs.
\newblock \emph{Advances in neural information processing systems}, 30.

\bibitem[{Hill(2012)}]{hill2012encyclopedia}
Hill, A.~V. 2012.
\newblock \emph{The encyclopedia of operations management: a field manual and glossary of operations management terms and concepts}.
\newblock Ft Press.

\bibitem[{Hodson(2022)}]{hodson2022root}
Hodson, T.~O. 2022.
\newblock Root-mean-square error (RMSE) or mean absolute error (MAE): When to use them or not.
\newblock \emph{Geoscientific Model Development}, 15(14): 5481--5487.

\bibitem[{Horn, Pack, and Rieger(2020)}]{horn2020autofeat}
Horn, F.; Pack, R.; and Rieger, M. 2020.
\newblock The autofeat python library for automated feature engineering and selection.
\newblock In \emph{Machine Learning and Knowledge Discovery in Databases: International Workshops of ECML PKDD 2019, W{\"u}rzburg, Germany, September 16--20, 2019, Proceedings, Part I}, 111--120. Springer.

\bibitem[{Hu et~al.(2023)Hu, Fan, Yi, Wang, Xu, Fu, and Wang}]{hu2023boosting}
Hu, X.; Fan, W.; Yi, K.; Wang, P.; Xu, Y.; Fu, Y.; and Wang, P. 2023.
\newblock Boosting Urban Prediction via Addressing Spatial-Temporal Distribution Shift.
\newblock In \emph{2023 IEEE International Conference on Data Mining (ICDM)}, 160--169. IEEE Computer Society.

\bibitem[{Hu et~al.(2024)Hu, Wang, Ying, and Fu}]{hu2024reinforcement}
Hu, X.; Wang, D.; Ying, W.; and Fu, Y. 2024.
\newblock Reinforcement Feature Transformation for Polymer Property Performance Prediction.
\newblock In \emph{Proceedings of the 33rd ACM International Conference on Information and Knowledge Management}, 4538--4545.

\bibitem[{Huo et~al.(2025{\natexlab{a}})Huo, Lu, Li, Zhu, and Chen}]{huo2025ct}
Huo, M.; Lu, K.; Li, Y.; Zhu, Q.; and Chen, Z. 2025{\natexlab{a}}.
\newblock Ct-patchtst: Channel-time patch time-series transformer for long-term renewable energy forecasting.
\newblock \emph{arXiv preprint arXiv:2501.08620}.

\bibitem[{Huo et~al.(2025{\natexlab{b}})Huo, Lu, Zhu, and Chen}]{huo2025enhancing}
Huo, M.; Lu, K.; Zhu, Q.; and Chen, Z. 2025{\natexlab{b}}.
\newblock Enhancing customer contact efficiency with graph neural networks in credit card fraud detection workflow.
\newblock \emph{arXiv preprint arXiv:2504.02275}.

\bibitem[{Izmailov et~al.(2018)Izmailov, Podoprikhin, Garipov, Vetrov, and Wilson}]{izmailov2018averaging}
Izmailov, P.; Podoprikhin, D.; Garipov, T.; Vetrov, D.; and Wilson, A.~G. 2018.
\newblock Averaging weights leads to wider optima and better generalization.
\newblock \emph{arXiv preprint arXiv:1803.05407}.

\bibitem[{Kanter and Veeramachaneni(2015)}]{kanter2015deep}
Kanter, J.~M.; and Veeramachaneni, K. 2015.
\newblock Deep feature synthesis: Towards automating data science endeavors.
\newblock In \emph{2015 IEEE international conference on data science and advanced analytics (DSAA)}, 1--10. IEEE.

\bibitem[{Khurana, Samulowitz, and Turaga(2018)}]{khurana2018feature}
Khurana, U.; Samulowitz, H.; and Turaga, D. 2018.
\newblock Feature engineering for predictive modeling using reinforcement learning.
\newblock In \emph{Proceedings of the AAAI Conference on Artificial Intelligence}, volume~32.

\bibitem[{Khurana et~al.(2016)Khurana, Turaga, Samulowitz, and Parthasrathy}]{khurana2016cognito}
Khurana, U.; Turaga, D.; Samulowitz, H.; and Parthasrathy, S. 2016.
\newblock Cognito: Automated feature engineering for supervised learning.
\newblock In \emph{2016 IEEE 16th International Conference on Data Mining Workshops (ICDMW)}, 1304--1307. IEEE.

\bibitem[{Kim et~al.(2021)Kim, Kim, Tae, Park, Choi, and Choo}]{kim2021reversible}
Kim, T.; Kim, J.; Tae, Y.; Park, C.; Choi, J.-H.; and Choo, J. 2021.
\newblock Reversible instance normalization for accurate time-series forecasting against distribution shift.
\newblock In \emph{International Conference on Learning Representations}.

\bibitem[{Kusiak(2001)}]{kusiak2001feature}
Kusiak, A. 2001.
\newblock Feature transformation methods in data mining.
\newblock \emph{IEEE Transactions on Electronics packaging manufacturing}, 24(3): 214--221.

\bibitem[{Li, Tang, and Li(2024)}]{li2024adaer}
Li, X.; Tang, B.; and Li, H. 2024.
\newblock AdaER: An adaptive experience replay approach for continual lifelong learning.
\newblock \emph{Neurocomputing}, 572: 127204.

\bibitem[{Ma{\'c}kiewicz and Ratajczak(1993)}]{mackiewicz1993principal}
Ma{\'c}kiewicz, A.; and Ratajczak, W. 1993.
\newblock Principal components analysis (PCA).
\newblock \emph{Computers \& Geosciences}, 19(3): 303--342.

\bibitem[{Shen et~al.(2020)Shen, Cui, Zhang, and Kunag}]{shen2020stable}
Shen, Z.; Cui, P.; Zhang, T.; and Kunag, K. 2020.
\newblock Stable learning via sample reweighting.
\newblock In \emph{Proceedings of the AAAI Conference on Artificial Intelligence}, volume~34, 5692--5699.

\bibitem[{Tran, Xue, and Zhang(2016)}]{tran2016genetic}
Tran, B.; Xue, B.; and Zhang, M. 2016.
\newblock Genetic programming for feature construction and selection in classification on high-dimensional data.
\newblock \emph{Memetic Computing}, 8: 3--15.

\bibitem[{Vanschoren et~al.(2013)Vanschoren, van Rijn, Bischl, and Torgo}]{OpenML2013}
Vanschoren, J.; van Rijn, J.~N.; Bischl, B.; and Torgo, L. 2013.
\newblock OpenML: networked science in machine learning.
\newblock \emph{SIGKDD Explorations}, 15(2): 49--60.

\bibitem[{Wang et~al.(2022)Wang, Fu, Liu, Li, and Solihin}]{wang2022group}
Wang, D.; Fu, Y.; Liu, K.; Li, X.; and Solihin, Y. 2022.
\newblock Group-wise reinforcement feature generation for optimal and explainable representation space reconstruction.
\newblock In \emph{Proceedings of the 28th ACM SIGKDD Conference on Knowledge Discovery and Data Mining}, 1826--1834.

\bibitem[{Wang et~al.(2023)Wang, Xiao, Wu, Wang, Zhou, and Fu}]{wang2023reinforcement}
Wang, D.; Xiao, M.; Wu, M.; Wang, P.; Zhou, Y.; and Fu, Y. 2023.
\newblock Reinforcement-Enhanced Autoregressive Feature Transformation: Gradient-steered Search in Continuous Space for Postfix Expressions.
\newblock \emph{arXiv preprint arXiv:2309.13618}.

\bibitem[{Wang et~al.(2025)Wang, Bai, Gong, Ying, Dong, Cui, and Fu}]{wang2025llm}
Wang, X.; Bai, H.; Gong, N.; Ying, W.; Dong, S.; Cui, X.; and Fu, Y. 2025.
\newblock LLM-ML Teaming: Integrated Symbolic Decoding and Gradient Search for Valid and Stable Generative Feature Transformation.
\newblock \emph{arXiv preprint arXiv:2506.09085}.

\bibitem[{Wang et~al.(2024)Wang, Wang, Ying, Xie, Chen, and Fu}]{wang2024knockoff}
Wang, X.; Wang, D.; Ying, W.; Xie, R.; Chen, H.; and Fu, Y. 2024.
\newblock Knockoff-Guided Feature Selection via A Single Pre-trained Reinforced Agent.
\newblock \emph{arXiv preprint arXiv:2403.04015}.

\bibitem[{Ying et~al.(2024{\natexlab{a}})Ying, Bai, Liu, and Fu}]{ying2024topology}
Ying, W.; Bai, H.; Liu, K.; and Fu, Y. 2024{\natexlab{a}}.
\newblock Topology-aware Reinforcement Feature Space Reconstruction for Graph Data.
\newblock \emph{arXiv preprint arXiv:2411.05742}.

\bibitem[{Ying et~al.(2024{\natexlab{b}})Ying, Wang, Chen, and Fu}]{ying2024feature}
Ying, W.; Wang, D.; Chen, H.; and Fu, Y. 2024{\natexlab{b}}.
\newblock Feature selection as deep sequential generative learning.
\newblock \emph{ACM Transactions on Knowledge Discovery from Data}, 18(9): 1--21.

\bibitem[{Ying et~al.(2024{\natexlab{c}})Ying, Wang, Hu, Qiu, Park, and Fu}]{ying2024revolutionizing}
Ying, W.; Wang, D.; Hu, X.; Qiu, J.; Park, J.; and Fu, Y. 2024{\natexlab{c}}.
\newblock Revolutionizing Biomarker Discovery: Leveraging Generative AI for Bio-Knowledge-Embedded Continuous Space Exploration.
\newblock In \emph{Proceedings of the 33rd ACM International Conference on Information and Knowledge Management}, 5046--5053.

\bibitem[{Ying et~al.(2024{\natexlab{d}})Ying, Wang, Hu, Zhou, Aggarwal, and Fu}]{ying2024unsupervised}
Ying, W.; Wang, D.; Hu, X.; Zhou, Y.; Aggarwal, C.~C.; and Fu, Y. 2024{\natexlab{d}}.
\newblock Unsupervised generative feature transformation via graph contrastive pre-training and multi-objective fine-tuning.
\newblock In \emph{Proceedings of the 30th ACM SIGKDD Conference on Knowledge Discovery and Data Mining}, 3966--3976.

\bibitem[{Ying et~al.(2023)Ying, Wang, Liu, Sun, and Fu}]{ying2023self}
Ying, W.; Wang, D.; Liu, K.; Sun, L.; and Fu, Y. 2023.
\newblock Self-optimizing feature generation via categorical hashing representation and hierarchical reinforcement crossing.
\newblock In \emph{2023 IEEE International Conference on Data Mining (ICDM)}, 748--757. IEEE.

\bibitem[{Ying et~al.(2025)Ying, Wei, Gong, Wang, Bai, Malarkkan, Dong, Wang, Zhang, and Fu}]{ying2025survey}
Ying, W.; Wei, C.; Gong, N.; Wang, X.; Bai, H.; Malarkkan, A.~V.; Dong, S.; Wang, D.; Zhang, D.; and Fu, Y. 2025.
\newblock A Survey on Data-Centric AI: Tabular Learning from Reinforcement Learning and Generative AI Perspective.
\newblock \emph{arXiv preprint arXiv:2502.08828v2}.

\bibitem[{Yu et~al.(2023)Yu, Liu, Yang, and Wang}]{yu2023distribution}
Yu, R.; Liu, S.; Yang, X.; and Wang, X. 2023.
\newblock Distribution Shift Inversion for Out-of-Distribution Prediction.
\newblock In \emph{Proceedings of the IEEE/CVF Conference on Computer Vision and Pattern Recognition}, 3592--3602.

\bibitem[{Zha et~al.(2023)Zha, Bhat, Lai, Yang, Jiang, Zhong, and Hu}]{zha2023data}
Zha, D.; Bhat, Z.~P.; Lai, K.-H.; Yang, F.; Jiang, Z.; Zhong, S.; and Hu, X. 2023.
\newblock Data-centric artificial intelligence: A survey.
\newblock \emph{arXiv preprint arXiv:2303.10158}.

\bibitem[{Zhang et~al.(2023)Zhang, Yan, Li, Wang, Xie, Song, and Kim}]{zhang2023continual}
Zhang, P.; Yan, Y.; Li, C.; Wang, S.; Xie, X.; Song, G.; and Kim, S. 2023.
\newblock Continual Learning on Dynamic Graphs via Parameter Isolation.
\newblock \emph{arXiv preprint arXiv:2305.13825}.

\bibitem[{Zhu et~al.(2025)Zhu, Lu, Huo, and Li}]{zhu2025image}
Zhu, Q.; Lu, K.; Huo, M.; and Li, Y. 2025.
\newblock Image-to-image translation with diffusion transformers and clip-based image conditioning.
\newblock \emph{arXiv preprint arXiv:2505.16001}.

\end{thebibliography}

\appendix

\section{RL Agents for Data Collection}
\label{appendix:RL}
\noindent\textbf{Leveraging reinforcement learning to explore high-quality and diverse training data.} Inspired by the exploitation, exploration, and self-learning abilities of reinforcement learning, our idea is to view an RL agent as a training data collector in order to achieve volume (self-learning enabled automation), diversity (exploration),  and quality (exploitation). Specifically, we design RL agents to automatically decide how to perform feature cross and generate new feature sets. \textbf{Figure~\ref{data_prepare}} shows that the reinforcement exploration experiences and corresponding accuracy will be collected and stored as training data. The training data collector includes:  
1) \noindent\underline{Multiple Agents: }  We design three agents to perform feature crossing: a head feature agent, an operation agent, and a tail feature agent. 
2)  \noindent\underline{Actions: }  In each reinforcement iteration, the three agents collaborate to select a head feature, an operator (e.g., +,-,*,/), and a tail feature to generate a new feature (e.g. $f_1$+$f_2$). The newly generated feature is later added to the feature set for the next feature generation. To balance diversity and quality, we employ the $\epsilon$-greedy DQN algorithm. For each step, each agent has a $\epsilon$ probability of selecting a feature/operation based on policy and has a (1-$\epsilon$) probability of selecting a feature/operation. The $\epsilon$ will increase over time so that agents can first explore more and then exploit more. 
3)  \noindent\underline{Environment: } The environment is the feature space, representing an updated feature set. When three feature agents generate and add a new feature to the previous feature set, the state of the feature space (i.e., the environment) changes. The state represents the statistical characteristics (such as aggregated average mean, and variance) of the feature subspace. 
4)  \noindent\underline{Reward: } We formulate the reward as the improvement of the performance of the explored feature set on a downstream task on the current iteration, compared with that of the explored feature set on the previous iteration. In this way, we reinforce agents to explore high-quality feature sets. 
5)  \noindent\underline{Training and Optimization: }  Our reinforcement data collector includes many feature crossing steps. Each step consists of two stages - control and training. In the control stage, each feature agent takes actions to change the size and contents of a new feature set and compute a reward. This reward is assigned to each participating agent. In the training stage, the agents train their policies via experience replay independently by minimizing the mean squared error (MSE) of the Bellman Equation. 
6)  \noindent\underline{Transformed Feature Set-Performance Pairs:} We introduce a hyperparameter to control the length of RL-exlored feature set and test each of them with a downstream ML task (e.g., random forest classification) to collect feature set-performance pairs. We select the top 5,000 feature set-performance pairs based on accuracy as training data.

\section{Overall framework}
\label{appendix:overall}
\setlength{\textfloatsep}{2pt}
\begin{algorithm}[h]
    \caption{Overall Framework}
    \label{alg_overall}
    \begin{algorithmic}[1]
    \REQUIRE Original dataset
    \ENSURE Optimal Transformed Feature set

    \STATE Initialize the encoder, evaluator, and decoder
    \STATE Leveraging RL agents to collect high-quality data
    \STATE Construct feature-feature similarity graphs
    \STATE Normalize the collected data
    \STATE Shift-resistant Bilevel Training to train encoder, \\
    decoder, and evaluator
    \STATE Identify the optimal embedding
    \STATE Leverage a well-trained decoder to generate the feature transformation operation sequences in an auto-\\
    regressive manner based on the optimal embedding
    \STATE Denormalize the Optimal Transformed Feature set
    \end{algorithmic}
\end{algorithm}

\textbf{Algorithm~\ref{alg_overall}} describes the overall process. 
It begins by initializing three core components: an encoder to represent features, an evaluator to assess transformations, and a decoder to generate transformation operations.
RL agents are employed to explore the transformation space and collect high-quality transformed data. These data are used to construct a feature-feature similarity graph, enabling the model to capture structural relationships among features. After normalization, a shift-resistant bilevel training strategy is applied to jointly train the encoder, decoder, and evaluator, ensuring robustness against data distribution shifts.
Once training stabilizes, the optimal embedding is selected. Based on this embedding, the decoder generates transformation operation sequences in an autoregressive manner, which are then used to produce the transformed features. Finally, a denormalization step is applied to recover the optimal transformed feature set in the original data space.

\section{Experimental Setup}
\label{appendix:exp_setup}
\noindent\textbf{Datasets.}
We perform experiments on 16 publicly available datasets from UCI~\cite{Dua:2019} and OpenML~\cite{OpenML2013}, which are widely used to verify tabular learning methods. 
To evaluate the robustness of \model\ against distribution shifts, we iteratively generate training and testing sets.
Starting from the first feature, we split the dataset into 80\% for training and 20\% for testing based on their observation index. Using the Kolmogorov-Smirnov test~\cite{berger2014kolmogorov} at a 95\% confidence level, we identify distribution shifts. If detected, we finalize the training and testing sets with allocations of 80\% and 20\%. If no shift is detected, we move to the next feature. If no shift is detected across all features, we exclude the dataset.

\noindent\textbf{Baselines.}
We compare our framework with 8 wildly-used feature transformation methods:
1) \textbf{Random Generation (RDG)} randomly produces feature-operation-feature transformations to generate a new feature space; 
2) \textbf{Essential Random Generation (ERG)} initially expands the feature space by applying operations to each feature, and then chooses the essential features as the new feature space; 
3) \textbf{Latent Dirichlet Allocation (LDA)}~\cite{blei2003latent} refines the feature space to get the factorized hidden state via matrix factorization; 
4) \textbf{AutoFeat Automated Transformation (AFAT)}~\cite{horn2020autofeat} is an enhanced version of ERG, which repeatedly generates new features and employs multi-step feature selection to identify informative ones; 
5) \textbf{Neural Feature Search (NFS)}~\cite{chen2019neural} generates the transformation sequence for each feature and the entire process is optimized by RL; 
6) \textbf{Traversal Transformation Graph (TTG)}~\cite{khurana2018feature} formulates the transformation process as a graph and subsequently employs an RL-based search method to find the optimal feature set; 
7) \textbf{Group-wise Reinforcement Feature Generation (GRFG)}~\cite{wang2022group} employs three collaborative reinforced agents to perform feature generation for feature space refinement.
8) \textbf{reinforceMent-enhanced autOregressive feAture Transformation (MOAT)}~\cite{wang2023reinforcement} formulates the discrete feature transformation problem as a continuous optimization task, while still assuming the I.I.D. condition.
9) \textbf{uNsupervised gEnerative feAture Transformation (NEAT)}~\cite{ying2024unsupervised}  introduces a contrastive learning framework that optimizes feature transformations without using labels, by aligning representations from original and transformed features through self-supervised objectives.
10). \textbf{Evolutionary Large Language Model framework for automated Feature Transformation (ELLM-FT)}~\cite{gong2025evolutionary} formulates feature transformation as a sequential generation task, leveraging few-shot prompting of LLMs guided by a reinforcement-evolved database to enable efficient and generalizable optimization beyond task-specific knowledge.

\noindent\textbf{Evaluation Metrics \& Hyperparameters Setting.}
To control the variance of the downstream model's impact on evaluation, we apply Random Forests (RF) for both classification and regression tasks. For classification, we use the F1-score, Precision, and Recall as the evaluation metrics. In regression, we assess performance using the following metrics ~\cite{hill2012encyclopedia,hodson2022root} 1 - Relative Absolute Error (1-RAE), 1 - Mean Absolute Error (1-MAE), and 1 - Mean Squared Error (1-MSE). For all these metrics, a higher value indicates a more effective feature transformation space.
1) \ul{RL collector:} We use the reinforcement data collector
to collect explored feature set-feature utility score pairs. The collector explores 512 episodes, and each episode includes 10 steps. 2) \ul{Graph Construction:} The threshold for creating an edge between two features is the 95th percentile of all similarity values. 3) \ul{Our framework:} we map the attribute of each node to a 64-dimensional embedding, and use a 2-layer GNN network and a 2-layer projection head to integrate such information. The decoder is a 1-layer LSTM network, which reconstructs a feature cross sequence. The evaluator is a 2-layer feed-forward network, in which the dimension of each layer is
200. During optimization, During optimization, we assign $\alpha$ (i.e., 10) as the weight for estimation loss and $\beta$ (i.e., 0.1) as the weight for reconstruction loss, the batch size is 256, the epochs are 500, and the learning rate range is 0.001-0.0005.

\noindent\textbf{Environmental Settings.}
Experiments are conducted on the Ubuntu 22.04.3 LTS operating system, Intel(R) Core(TM) i9-13900KF CPU@ 3GHz, and 1 way RTX 4090 and 32GB of RAM, with the framework of Python 3.11.4 and PyTorch 2.0.1.

\subsection{Qualitative Analysis}
\label{appendix:qualitative}
\begin{figure}[]
	\centering
	\includegraphics[width=0.95\linewidth]{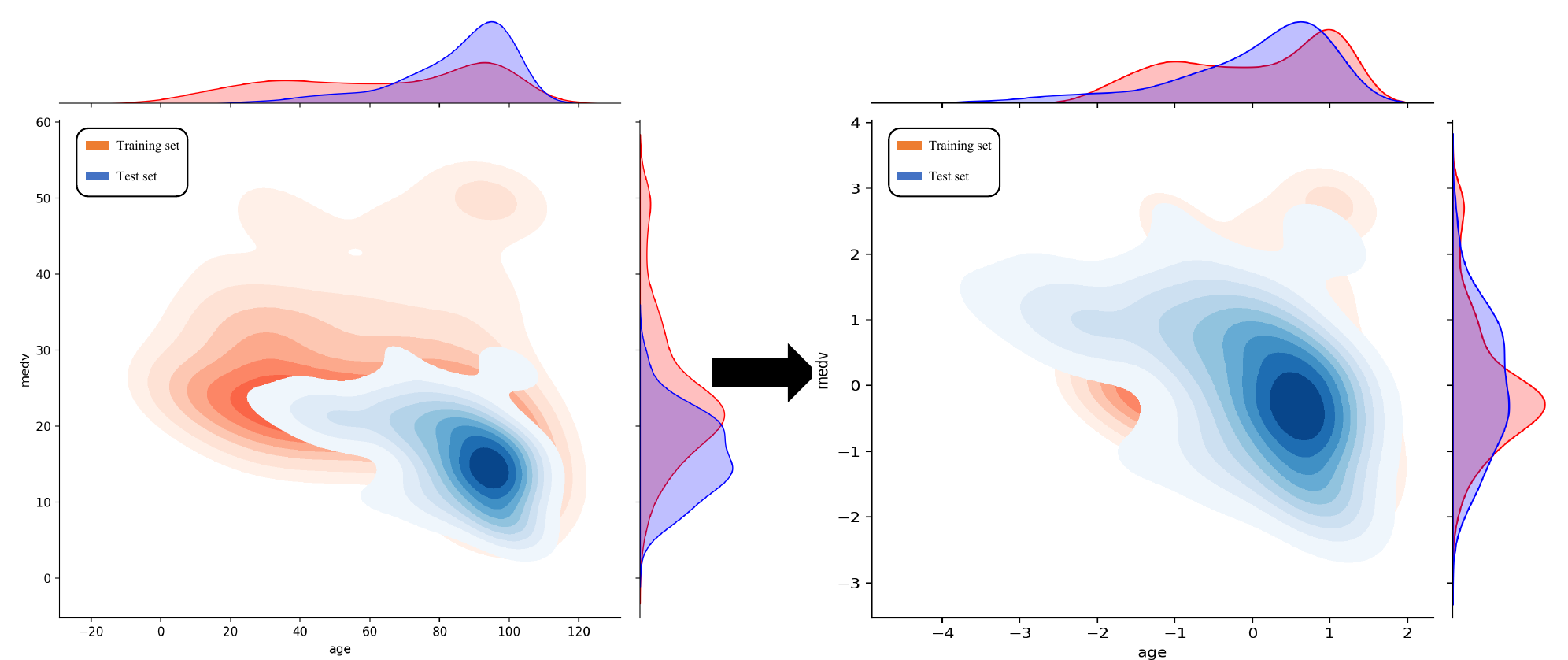}
	\caption{Changes in the joint and marginal distributions of age and MEDV (target) on Housing Boston, comparing datasets before (left) and after (right) normalization.}
    \label{normhb}
\end{figure}

\begin{figure}[]
	\centering
	\includegraphics[width=0.95\linewidth]{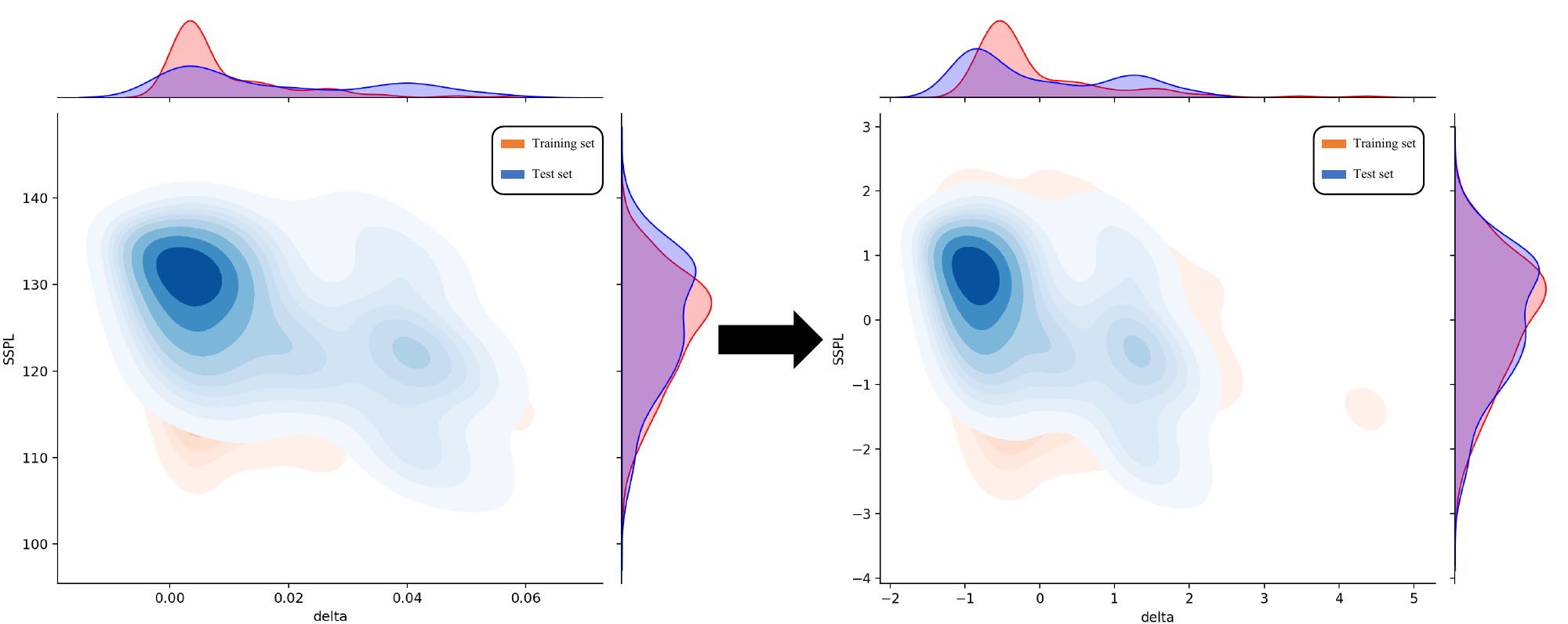}
	\caption{Comparison of joint and marginal distributions of delta and SSPL (target) on the Airfoil dataset before (left) and after (right) normalization, for both training and test sets.}
    \label{normair}
\end{figure}

\noindent\textbf{Shift Elimination.}
We conduct a case study to examine the effect of normalization in reducing distribution shifts between training and test sets. \textbf{Figures~\ref{normhb}} and \textbf{\ref{normair}} show the distributional changes in the Housing Boston and Airfoil datasets, respectively. After normalization, the joint and marginal distributions of age and MEDV (target) become significantly aligned across the training and test sets. In the Airfoil dataset, while the joint distribution of delta and SSPL (target) remains relatively stable, the marginal distributions are better aligned. These results confirm that normalization effectively mitigates distribution shift between training and test data.

\setlength{\tabcolsep}{1.6mm}{
\begin{table*}[t]
\caption{Time cost comparisons with baselines.}
\centering
\fontsize{7}{6}\selectfont
\begin{tabular}{@{}ccccccccccc@{}}
\toprule
Dataset            & RL Data Collection (min) & RDG(s) & LDA(s) & ERG(s) & AFAT(s) & NFS(s) & TTG(s) & GRFG(s) & MOAT(min) & \model(min) \\ \midrule
SpectF             & 33.7                   & 3.0    & 0.3    & 9.5    & 3.8     & 8.3    & 9.1    & 282.1   & 156.0     & 142.1        \\
Wine Quality White & 114.7                  & 30.8   & 0.5    & 29.4   & 4.2     & 24.7   & 26.4   & 373.2   & 195.8     & 39.9        \\
Wine Quality Red   & 36.5                   & 7.4    & 0.6    & 13.9   & 3.6     & 9.2    & 10.3   & 182.2   & 107.5     & 43.1        \\
openml\_616        & 156.6                  & 27.5   & 0.2    & 55.9   & 2.6     & 18.1   & 24.6   & 936.5   & 430.1     & 160.7        \\
openml\_618        & 239.6                  & 67.7   & 0.2    & 82.4   & 3.0     & 30.9   & 38.7   & 821.6   & 373.8     & 95.9        \\
openml\_637        & 171.3                  & 28.1   & 0.2    & 55.8   & 2.8     & 19.3   & 25.1   & 731.7   & 300.9     & 147.9        \\ \bottomrule
\end{tabular}
\label{time_cost}
\end{table*}}
\setlength{\tabcolsep}{0.75mm}{
\begin{table*}[t]
\caption{Space complexity comparisons with baselines.}
\centering
\fontsize{7}{6}\selectfont
\begin{tabular}{@{}ccccccccccc@{}}
\toprule
Dataset            & RL Data Collection (MB) & RDG(MB) & LDA(MB) & ERG(MB) & AFAT(MB) & NFS(MB) & TTG(MB) & GRFG(MB) & MOAT(MB) & \model(MB) \\ \midrule
SpectF             & 0.19                  & 0.13    & 0.07    & 0.14    & 0.08     & 0.14    & 0.13    & 1.04     & 0.14     & 0.16      \\
Wine Quality White & 0.21                  & 0.15    & 0.08    & 0.16    & 0.09     & 0.15    & 0.14    & 18.70     & 0.13     & 0.48      \\
Wine Quality Red   & 0.19                  & 0.14    & 0.07    & 0.15    & 0.08     & 0.15    & 0.15    & 6.14     & 0.13     & 0.19      \\
openml\_616        & 0.20                  & 0.14    & 0.07    & 0.16    & 0.08     & 0.16    & 0.14    & 1.94     & 0.14     & 0.17      \\
openml\_618        & 0.23                  & 0.15    & 0.09    & 0.16    & 0.11     & 0.17    & 0.14    & 3.84     & 0.14     & 0.23      \\
openml\_637        & 0.23                  & 0.16    & 0.10    & 0.17    & 0.11     & 0.17    & 0.15    & 1.94     & 0.14     & 0.17      \\ \bottomrule
\end{tabular}
\vspace{-0.4cm}
\label{space_cost}
\end{table*}}


\begin{figure}[]
\vspace{-0.1cm}
        \centering
        \subfigure[Training Set]{\label{training}\includegraphics[width=0.15\textwidth]{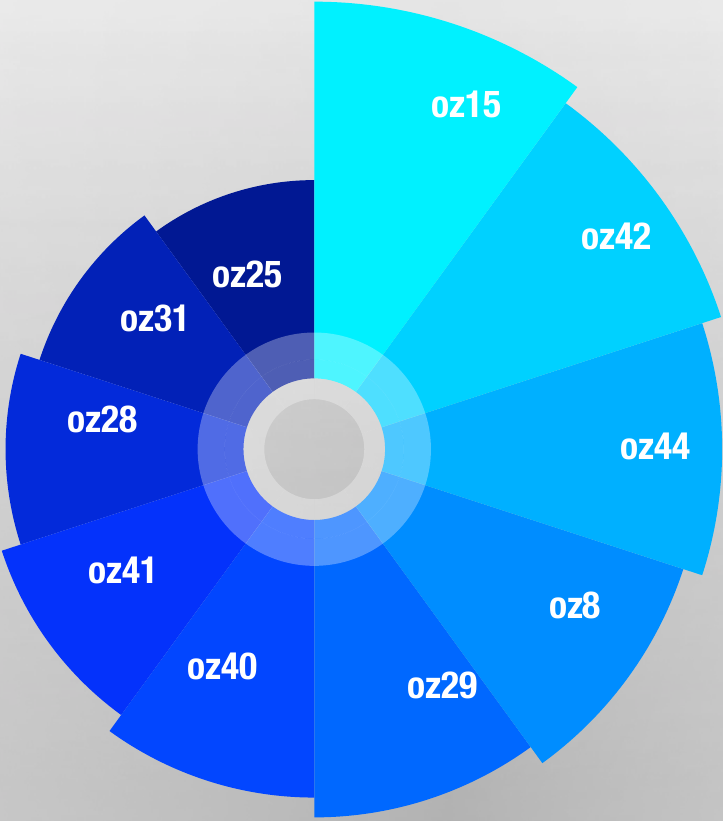}}
        \subfigure[Test Set]{\label{training}\includegraphics[width=0.15\textwidth]{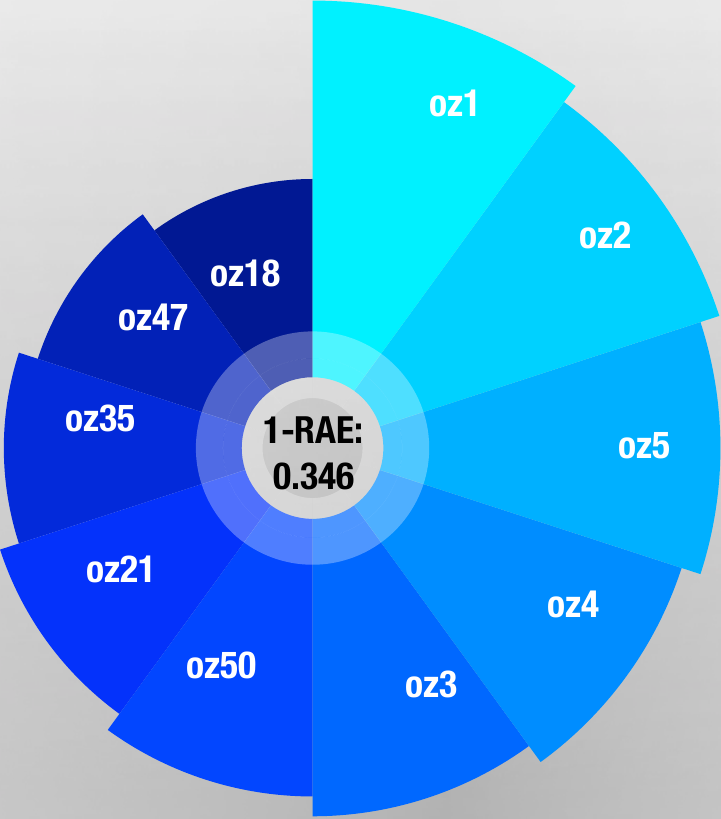}}
        \subfigure[\model\ Generated Feature Set]{\label{ours}\includegraphics[width=0.15\textwidth, height=0.17\textwidth]{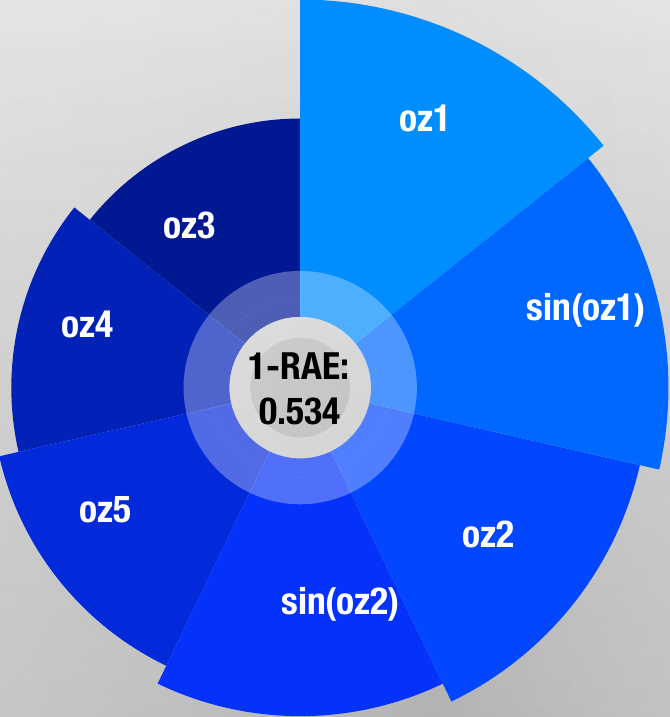}}
        \caption{Comparison of the most important features in the original feature space of training and test set, and the \model\ generated feature space.}
        \label{feature importance}
\end{figure}

\noindent\textbf{Feature Importance.}
We select the top 10 most important features—measured by mutual information with the target—from the original training set, test set, and the transformed feature space generated by \model\ on the openml\_616 dataset.
\textbf{Figure~\ref{feature importance}} shows that important features differ significantly between training and test sets, highlighting the impact of distribution shift and the unreliability of directly transferring feature importance across domains.
Notably, \model\ selects only 7 features (including 2 newly generated ones), yet improves ML model performance by 29.24\%.
This demonstrates that \model\ can generate effective and transferable features by capturing robust and shift-invariant representations.

\subsection{Time and Space complexity comparisons with baselines}
\label{appendix:complexity_base}
We compare the computational cost of \model\ with baseline methods across six datasets. \textbf{Table~\ref{time_cost}} shows methods like RDG, LDA, NSF, and TTG are faster, but they perform worse due to their inability to handle distribution shifts. In contrast, \model\ achieves better accuracy with comparable or lower cost than GRFG and MOAT.
Notably, feature transformation is not a major bottleneck in most pipelines, and our method offers significant time savings over manual feature engineering.
While RL-based data collection introduces overhead, it is conducted offline and asynchronously. The additional cost stems from collecting high-quality data and training sequence models over complex feature operations.
\textbf{Table~\ref{space_cost}} shows \model\ maintains a small and stable memory footprint across datasets. This is enabled by the EOG framework, which embeds variable-length feature sequences into fixed-size vectors, thereby keeping the parameter size constant as the data grows. These results indicate that \model scales well both in time and space, maintains competitive efficiency, and ensures scalability for large-scale applications.

\begin{figure}[h]
\vspace{-0.3cm}
        \centering
        \subfigure[Model Parameter Size]{\label{training}\includegraphics[width=0.23\textwidth]{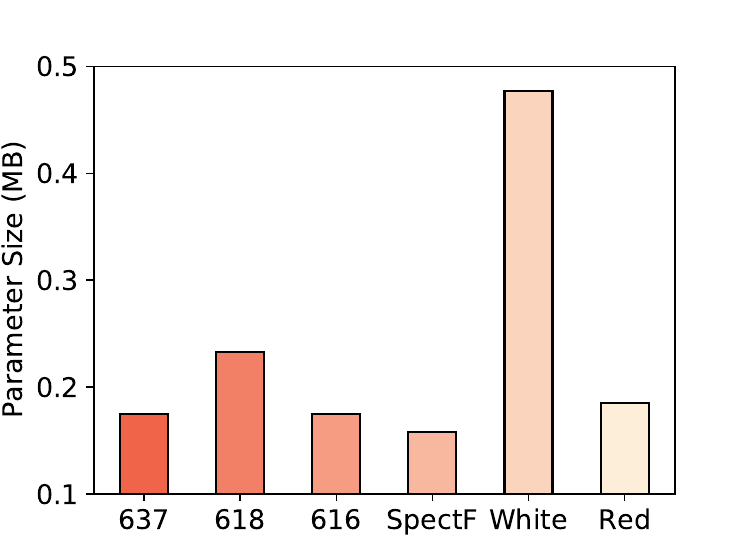}}
        \subfigure[Time Complexity]{\label{test}\includegraphics[width=0.23\textwidth]{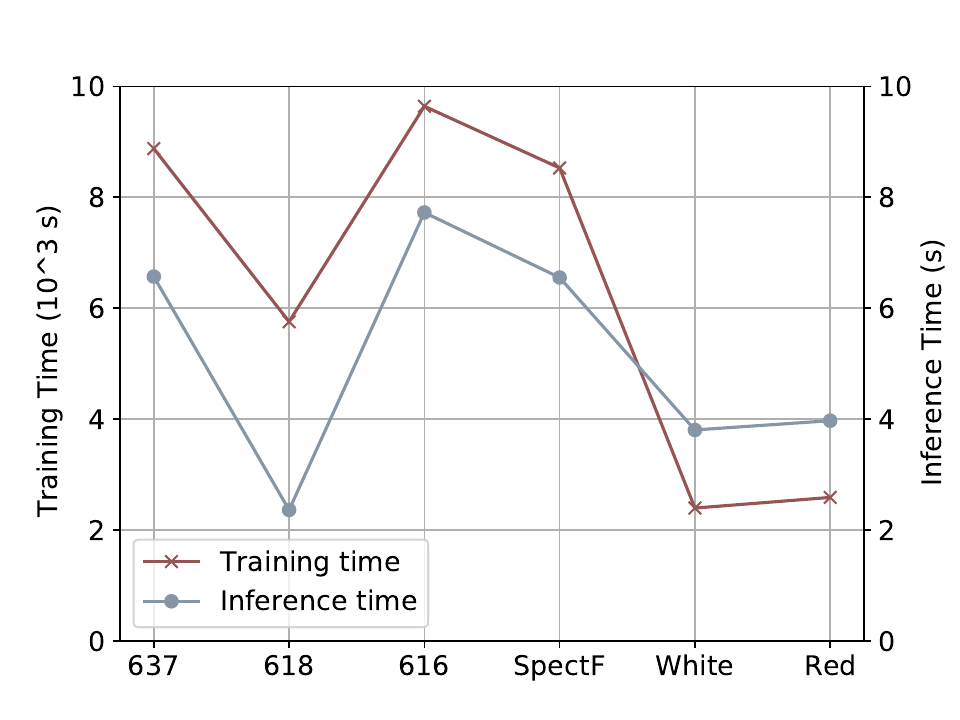}}
        \vspace{-0.2cm}
        \caption{Complexity analysis results (a) model parameter size and (b) Time Complexity.}
        \label{complexity}
\end{figure}
\subsection{Complexity Analysis of \model}
\label{appendix:complexity}
We investigate the space and time complexity, by comparing model parameter size, training time, and inference time across six datasets. \textbf{Figure~\ref{complexity}} demonstrates the results.
We can observe that the training and inference times for openml\_616, openml\_618, openml\_637, and SpectF are significantly higher than Wine Quality Red and Wine Quality White. The main reason is that, with an increasing number of features, the model has to exert more effort to capture the relevant representation, construct a robust embedding space, and generate the optimal transformed feature set.
However, the inference time is relatively low. Therefore, after the convergence of \model, we can quickly adapt to the unseen dataset with different distributions, making it practical for dynamic real-world scenarios. Moreover, \model\ upholds a small parameter size. The potential reason is that our GNN-based encoder can sample the most essential knowledge of the feature space to enhance the scalability and robustness of \model.

\end{document}